\PassOptionsToPackage{numbers,sort&compress}{natbib}
\documentclass[preprint,12pt]{elsarticle}

\usepackage{amssymb}
\usepackage{amsmath}

\usepackage{graphicx}
\usepackage{lineno}
\usepackage{array}
\usepackage{booktabs}
\usepackage{hyperref}
\usepackage{float} 
\usepackage{xcolor}
\usepackage{algorithm}
\usepackage{algpseudocode}
\usepackage{subcaption}
\usepackage{adjustbox}
\graphicspath{ {./f} }

\journal{Nuclear Engineering and Design}

\begin{document}


\begin{frontmatter}




\title{A Numerical Method for Coupling Parameterized Physics-Informed Neural Networks and FDM for Advanced Thermal-Hydraulic System Simulation}

\author[1]{Jeesuk Shin}

\author[2]{Donggyun Seo}

\author[3]{Sihyeong Yu}

\author[1]{Joongoo Jeon\corref{cor1}}
\ead{jgjeon41@postech.ac.kr}

\cortext[cor1]{Corresponding author}

\affiliation[1]{
  organization={Division of Advanced Nuclear Engineering, Pohang University of Science and Technology},
  city={Pohang-si},
  country={Republic of Korea}
}

\affiliation[2]{
  organization={Mueunjae School of Undergraduate Studies, Pohang University of Science and Technology},
  city={Pohang-si},
  country={Republic of Korea}
}

\affiliation[3]{
  organization={Department of Quantum System Engineering, Jeonbuk National University},
  city={Jeonju-si},
  country={Republic of Korea}
}

\begin{abstract}
Severe accident analysis using system-level codes such as MELCOR is indispensable for nuclear safety assessment, yet the computational cost of repeated simulations poses a significant bottleneck for parametric studies and uncertainty quantification.
Existing surrogate models accelerate these analyses but depend on large volumes of simulation data, while physics-informed neural networks (PINNs) enable data-free training but must be retrained for every change in problem parameters.
This study addresses both limitations by developing the Parameterized PINNs coupled with FDM (P2F) method, a node-assigned hybrid framework for MELCOR's Control Volume Hydrodynamics/Flow Path (CVH/FP) module.
In the P2F method, a parameterized Node-Assigned PINN (NA-PINN) accepts the water-level difference, initial velocity, and time as inputs, learning a solution manifold so that a single trained network serves as a data-free surrogate for the momentum conservation equation across all flow paths without retraining.
This PINN is coupled with an FDM solver that advances the mass conservation equation at each time step, ensuring exact discrete mass conservation while replacing the iterative nonlinear momentum solve with a single forward pass.
Verification on a six-tank gravity-driven draining scenario yields a water level mean absolute error of $7.85 \times 10^{-5}$~m and a velocity mean absolute error of $3.21 \times 10^{-3}$~m/s under the nominal condition with $\Delta t = 1.0$~s.
The framework maintains consistent accuracy across time steps from 0.2 to 1.0~s and generalizes to five distinct initial conditions, all without retraining or simulation data.
This work introduces numerical method for coupling parameterized PINN and FDM within a nuclear thermal-hydraulic system code.
\end{abstract}



\begin{keyword}
Physics-informed neural networks \sep Finite difference method \sep Thermal-hydraulics \sep MELCOR \sep Surrogate model \sep Coupling numerical method
\end{keyword}

\end{frontmatter}



\section{Introduction}\label{sec:1}
 
Severe accident (SA) analysis is essential for ensuring the safety of nuclear power plants, as such events can lead to significant consequences for both public health and the environment.
To simulate the progression of severe accidents and evaluate their potential impacts, system-level codes such as RELAP5/SCDAP, MAAP, and MELCOR have been widely employed \cite{iaea2019}.
Among these, MELCOR, developed by Sandia National Laboratories (SNL), adopts a control volume (CV) framework to model tightly coupled thermal-hydraulic and structural interactions spanning the entire spectrum of severe accident progression \cite{humphries2017, wagner2022}.
Despite its broad applicability, MELCOR presents substantial challenges arising from the complex multi-physics coupling among its constituent packages---particularly the CVH/FP and RadioNuclide (RN) packages ---which can compromise the reliability and stability of numerical simulations.
Moreover, when parametric studies, uncertainty quantification, or sensitivity analyses require a large number of simulation runs, the computational cost of repeated MELCOR executions becomes a significant practical bottleneck.
 
To address this challenge, artificial intelligence and machine learning techniques have been increasingly adopted across computational science and engineering to accelerate expensive numerical simulations, including neural network architectures that embed numerical discretization schemes~\cite{jeon2022finite, khanal2025comparison}.
Within the nuclear domain, data-driven surrogate models have emerged as a particularly prominent approach: artificial neural networks trained on datasets generated by MELCOR or similar codes have been developed to emulate simulation outputs at substantially reduced computational cost~\cite{chae2023, wangma2025, wangma2023, songha2022, lee2024surrogate, song2025dose}, and physics-informed surrogates incorporating physical constraints have also been explored~\cite{baraldi2025}.
While these surrogates offer rapid inference and parametric flexibility, they share a common dependence on large volumes of training data obtained from expensive code executions.

In contrast, physics-informed neural networks (PINNs)~\cite{raissi2019, karniadakis2021} offer a data-free alternative by embedding governing equations directly into the neural network loss function, enabling the solution of partial differential equations without reliance on external training data.
PINNs have been successfully demonstrated in heat transfer~\cite{cai2021heat, zhao2025}, fluid dynamics~\cite{cai2021fluid}, and the nuclear engineering domain~\cite{elhareef2023, ibarra2025}.
Despite these advances, a trained PINN is valid only for a single fixed set of initial and boundary conditions; changing any parameter requires complete retraining, precluding its use as a reusable surrogate model.

These complementary strengths and weaknesses of the two paradigms motivate the first methodological direction of this work.
Data-driven surrogates achieve parametric flexibility but require large volumes of expensive simulation data; PINNs eliminate the data requirement but sacrifice parametric generality.
A framework that combines data-free training with surrogate-level parametric flexibility---retaining the strengths of both while avoiding their respective limitations---has not yet been realized for thermal-hydraulic system code modeling.
 
Beyond the parametric limitation, PINNs face a second practical challenge: they are known to suffer from error growth in long-horizon time integration, where prediction errors accumulate over extended simulation domains~\cite{wang2024respecting, chen2024pinn}.
This limitation is particularly critical for thermal-hydraulic system codes, where simulations of severe accident progression routinely span hours to days of physical time, making robust long-term prediction an essential requirement.
A similar challenge has been observed in the CFD domain, where standalone ML-based surrogates suffer from error accumulation and instability over long simulations, motivating hybrid strategies that couple conventional numerical solvers with deep learning frameworks to restore robustness while retaining computational efficiency~\cite{jeon2024residual, baral2025residual}.
Motivated by a similar consideration, the present work couples a PINN surrogate with a conventional FDM solver within a shared time-marching loop: the PINN is assigned to selected physical nodes while the FDM solver handles the remaining nodes at every time step, so that the PINN only needs to predict over a single short time step rather than integrating over a long horizon, thereby inherently avoiding error accumulation.
No prior study has realized such a node-assigned hybrid coupling in which a data-free PINN surrogate and a conventional numerical solver are assigned to distinct physical nodes and advance together within a time-marching loop.
 
In our prior work~\cite{shin2025napinn}, we developed a Node-assigned PINN (NA-PINN) tailored to MELCOR's CVH/FP and HS packages, demonstrating the feasibility of a fully data-free, physics-informed approach to the lumped-parameter multi-physics structure of MELCOR.
However, NA-PINN---like standard PINNs---remains confined to a single fixed scenario, requiring complete retraining for any change in problem parameters, and is not coupled with a numerical solver to ensure long-horizon stability.
Both of the aforementioned gaps therefore remain unaddressed.
 
The present study aims to bridge both gaps by developing the Parameterized PINNs coupled with FDM (P2F) method, extending the NA-PINN concept to a parameterized setting inspired by the broader paradigm of parameterized PINNs \cite{cho2024, xie2024}
The proposed framework is verified on a six-tank gravity-driven draining scenario consistent with MELCOR's CVH/FP package formulation, demonstrating prediction accuracy comparable to the original NA-PINN, robustness to time-step variations, and generalization across diverse initial conditions without retraining.
The main contributions of this work are twofold:
\begin{enumerate}
    \item \textbf{The first data-free surrogate model for a nuclear thermal-hydraulic system code.}
    A parameterized PINN is developed that accepts physical state variables as additional inputs alongside the temporal coordinate, learning a solution manifold over the parameter space.
    In the present work, this network is trained once and applied across all flow path (FP) nodes, enabling predictions across a range of initial conditions without retraining or simulation data.
    \item \textbf{A node-assigned hybrid coupling strategy (P2F method).}
    The parameterized PINN handles the nonlinear momentum equation via a single forward pass, while an FDM solver advances the mass conservation equation at each time step.
    This strategic assignment is designed to exploit the strengths of each solver, ultimately aiming for both computational efficiency and prediction accuracy while preventing error accumulation through their alternating execution.
\end{enumerate}
 
The remainder of this paper is organized as follows.
Section~\ref{sec:2} provides the theoretical background, covering PINNs, the original NA-PINN architecture, parameterized PINNs, and the specific research gaps addressed in this study.
Section~\ref{sec:3} presents the six-tank problem setup and the training methodology for the parameterized PINN.
Section~\ref{sec:4} introduces the P2F framework and presents verification results.
Finally, Section~\ref{sec:6} concludes the paper and discusses future directions.

\section{Background}\label{sec:2}

\subsection{Physics-Informed Neural Networks}\label{sec:2.1}
 
Physics-informed neural networks (PINNs), introduced by Raissi et al.\ \cite{raissi2019}, approximate the solution of a partial differential equation (PDE) by training a neural network $u_\mathrm{NN}(\mathbf{x}, t; \boldsymbol{\theta})$ to satisfy the governing equation in its residual form.
For a general PDE $\mathcal{N}[u] = 0$, the PINN loss function penalizes the mean-squared residual evaluated at a set of collocation points, optionally augmented with boundary and initial condition terms.
All spatial and temporal derivatives are computed via automatic differentiation, eliminating the need for a computational mesh.
A key advantage over data-driven approaches is that no simulation data is required for training; the governing equations themselves serve as the sole source of supervision.
PINNs have been applied across heat transfer \cite{cai2021heat, zhao2025}, fluid dynamics \cite{cai2021fluid, arthurs2021}, structural and solid mechanics \cite{jeong2023, wang2023solid, jo2026task}, and other disciplines, in both data-free and data-assisted settings; a broader overview can be found in \cite{karniadakis2021}.
 
The general PINN training procedure is summarized in Algorithm~\ref{alg:general_pinn}.
Given a PDE $\mathcal{N}[u] = 0$ defined on a domain $\Omega$ with boundary conditions $\mathcal{B}[u] = 0$ on $\partial\Omega$ and an initial condition $u(\mathbf{x}, 0) = u_0(\mathbf{x})$, the network $u_\mathrm{NN}(\mathbf{x}, t; \boldsymbol{\theta})$ is trained by minimizing a composite loss function that penalizes the PDE residual, boundary condition violations, and initial condition discrepancies simultaneously.
 
\begin{algorithm}[H]
\caption{General Training Procedure of Physics-Informed Neural Networks}
\label{alg:general_pinn}
\begin{algorithmic}[1]
\Require PDE $\mathcal{N}[u] = 0$ on domain $\Omega$, boundary condition $\mathcal{B}[u] = 0$ on $\partial\Omega$, initial condition $u_0(\mathbf{x})$
\Require Collocation points: $N_r$ interior points $\{(\mathbf{x}_i^r,\, t_i^r)\}$, $N_b$ boundary points $\{(\mathbf{x}_i^b,\, t_i^b)\}$, $N_0$ initial points $\{\mathbf{x}_i^0\}$
\Require Number of epochs $N_{\mathrm{epoch}}$, learning rate $\eta$, loss weights $\lambda_r,\, \lambda_b,\, \lambda_0$
 
\State Initialize network parameters $\boldsymbol{\theta}$
 
\For{epoch $= 1, \dots, N_{\mathrm{epoch}}$}
 
    \Statex \hspace{\algorithmicindent}\hspace{\algorithmicindent} \textit{// Evaluate residuals via automatic differentiation}
    \State PDE residual: $\mathcal{L}_r = \dfrac{1}{N_r} \displaystyle\sum_{i=1}^{N_r} \left| \mathcal{N}[u_\mathrm{NN}](\mathbf{x}_i^r,\, t_i^r) \right|^2$
 
    \State Boundary loss: $\mathcal{L}_b = \dfrac{1}{N_b} \displaystyle\sum_{i=1}^{N_b} \left| \mathcal{B}[u_\mathrm{NN}](\mathbf{x}_i^b,\, t_i^b) \right|^2$
 
    \State Initial condition loss: $\mathcal{L}_0 = \dfrac{1}{N_0} \displaystyle\sum_{i=1}^{N_0} \left| u_\mathrm{NN}(\mathbf{x}_i^0,\, 0) - u_0(\mathbf{x}_i^0) \right|^2$
 
    \Statex \hspace{\algorithmicindent}\hspace{\algorithmicindent} \textit{// Compute total loss}
    \State $\mathcal{L}_{\mathrm{total}} = \lambda_r \, \mathcal{L}_r + \lambda_b \, \mathcal{L}_b + \lambda_0 \, \mathcal{L}_0$
 
    \Statex \hspace{\algorithmicindent}\hspace{\algorithmicindent} \textit{// Update parameters}
    \State $\boldsymbol{\theta} \leftarrow \boldsymbol{\theta} - \eta \, \nabla_{\boldsymbol{\theta}} \mathcal{L}_{\mathrm{total}}$
 
\EndFor
\State \Return trained model $u_\mathrm{NN}(\mathbf{x}, t; \boldsymbol{\theta})$
\end{algorithmic}
\end{algorithm}
 
\noindent A key feature of Algorithm~\ref{alg:general_pinn} is that all derivative terms required to evaluate $\mathcal{N}[u_\mathrm{NN}]$ are computed through automatic differentiation of the network output with respect to its inputs, requiring no mesh or discretization scheme. The loss weights $\lambda_r$, $\lambda_b$, and $\lambda_0$ balance the relative contributions of each term; however, appropriate tuning of these weights is often necessary, as improper balancing can lead to training difficulties~\cite{wang2021ntk}.
 
Within the nuclear engineering domain, Elhareef and Wu \cite{elhareef2023} demonstrated the feasibility of PINNs for solving neutron diffusion equations, while Coppo Leite et al.\ \cite{ibarra2025} applied a physics-informed convolutional neural network to monitor temperature fields in high-temperature gas reactors.
Despite these advances, the application of PINNs to lumped-parameter thermal-hydraulic system codes remained unexplored until our recent work on NA-PINN \cite{shin2025napinn}, described in the following subsection.
 
A well-documented challenge in applying PINNs to coupled multi-physics systems is the gradient conflict problem: when a single shared network learns multiple PDEs of different physical scales, competing gradient signals can impede convergence \cite{wang2021ntk}.
Strategies to mitigate this include adaptive loss weighting \cite{wang2021ntk}, domain decomposition with parallel subnetworks \cite{shukla2021, jagtap2020xpinn}, and multistage training \cite{khadijeh2025}.
The NA-PINN architecture addresses this challenge through a node-assignment strategy specifically tailored to MELCOR's CV structure.
 
\subsection{Node-Assigned PINN}\label{sec:2.2}
 
When a conventional shared-network PINN---taking time $t$ as input and simultaneously outputting all nodal variables through a single fully-connected network---was applied to the MELCOR CVH/FP equations, it failed to produce physically meaningful predictions for any tested configuration ($N_i = 2$, 3, and 6), converging instead to trivial constant-state solutions.
This failure is attributable to the severe scale disparity between the target variables: CV water heights (order of meters) and FP velocities (order of m/s) differ by magnitudes that cause the momentum residual gradients to dominate the loss landscape, effectively suppressing learning of the mass conservation equation.
 
\begin{figure}[H]
    \centering
    \includegraphics[width=0.90\linewidth]{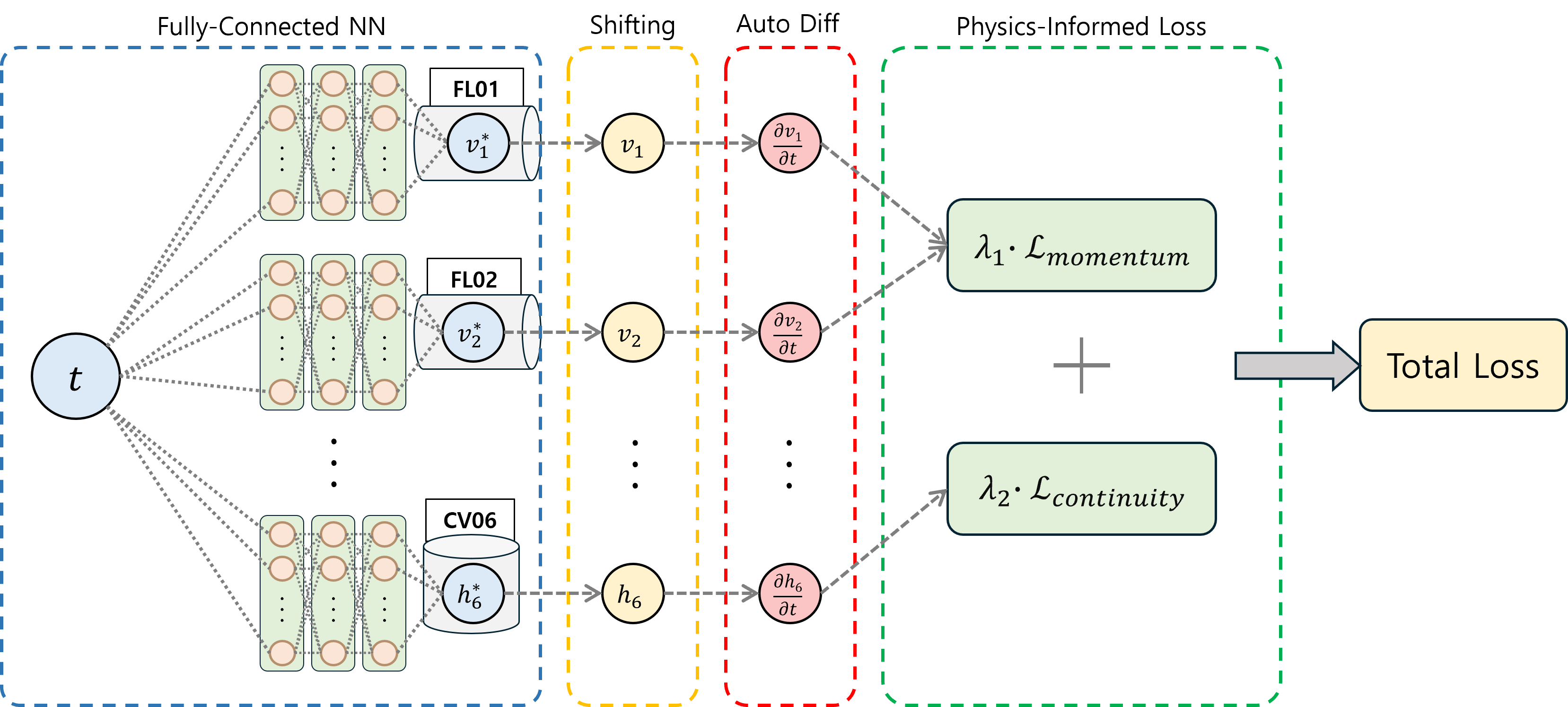}
    \caption{Architecture of the original NA-PINN~\cite{shin2025napinn}. Each nodal variable ($v_1, v_2, \ldots$ for FP velocities; $h_6$ for CV water heights) is predicted by a dedicated subnetwork that takes time $t$ as input. Temporal derivatives are computed via automatic differentiation, and the weighted momentum and continuity residuals are summed to form the total loss.}
    \label{fig:napinn_original_architecture}
\end{figure}

NA-PINN resolves this by assigning a dedicated subnetwork to each nodal variable---one for each CV water height and one for each FP velocity---while optimizing all subnetworks jointly through a unified physics-informed loss function that combines the momentum and continuity residuals.
This design eliminates gradient competition between variables of different physical scales while preserving inter-node coupling through the shared governing equations.
Figure~\ref{fig:napinn_original_architecture} illustrates the overall architecture: each subnetwork takes only the time coordinate $t$ as input and produces a single nodal output (either a velocity $v_j$ or a water height $h_i$), temporal derivatives are computed via automatic differentiation, and the weighted momentum ($\lambda_1 \cdot \mathcal{L}_{\mathrm{momentum}}$) and continuity ($\lambda_2 \cdot \mathcal{L}_{\mathrm{continuity}}$) residuals are summed to form the total loss.
 
The training procedure is formalized in Algorithm~\ref{alg:napinn_original}.
For a system of $n$ cascading tanks, the NA-PINN comprises $(n-1)$ velocity subnetworks and $n$ water-height subnetworks, each independently parameterized but jointly optimized through a shared loss function.
 
\begin{algorithm}[H]
\caption{Training Procedure of the Original NA-PINN~\cite{shin2025napinn}}
\label{alg:napinn_original}
\begin{algorithmic}[1]
\Require System of coupled PDEs $\{\mathcal{N}_k[u_1, \dots, u_N] = 0\}_{k=1}^{K}$ over $N$ spatial nodes
\Require Initial conditions $\{u_{0,i}\}_{i=1}^{N}$, collocation points $\{t_i\}_{i=1}^{N_t}$ on $[0,\, T]$
\Require Independent parameter sets $\boldsymbol{\Theta} = \{\boldsymbol{\theta}_1, \dots, \boldsymbol{\theta}_N\}$ (one per node)
\Require Number of epochs $N_{\mathrm{epoch}}$, learning rate scheduler $\eta_{\mathrm{sched}}$, loss weights $\{\lambda_k\}_{k=1}^{K}$
 
\State Initialize all subnetwork parameters $\boldsymbol{\Theta}$
 
\Statex \hspace{\algorithmicindent} \textit{// Enforce initial conditions via shifting (hard constraint)}
\State $\hat{u}_i(t) = \mathrm{NN}_{\boldsymbol{\theta}_i}(t) - \mathrm{NN}_{\boldsymbol{\theta}_i}(0) + u_{0,i}$ \quad for each node $i = 1, \dots, N$
 
\For{epoch $= 1, \dots, N_{\mathrm{epoch}}$}
 
    \Statex \hspace{\algorithmicindent}\hspace{\algorithmicindent} \textit{// Predict all node outputs}
    \State $\{\hat{u}_i(t)\}_{i=1}^{N} \leftarrow$ forward pass through each subnetwork
 
    \Statex \hspace{\algorithmicindent}\hspace{\algorithmicindent} \textit{// Compute temporal derivatives via automatic differentiation}
    \State $\left\{\dfrac{\partial \hat{u}_i}{\partial t}\right\}_{i=1}^{N}$
 
    \Statex \hspace{\algorithmicindent}\hspace{\algorithmicindent} \textit{// Evaluate PDE residuals for each equation $k$}
    \State $\mathcal{R}_k = \mathcal{N}_k[\hat{u}_1, \dots, \hat{u}_N]$ \quad for $k = 1, \dots, K$
 
    \Statex \hspace{\algorithmicindent}\hspace{\algorithmicindent} \textit{// Compute total loss (shared across all nodes)}
    \State $\mathcal{L}_{\mathrm{total}} = \displaystyle\sum_{k=1}^{K} \lambda_k \cdot \dfrac{1}{N_t} \sum_{i=1}^{N_t} \mathcal{R}_{k,i}^{\,2}$
 
    \Statex \hspace{\algorithmicindent}\hspace{\algorithmicindent} \textit{// Update all subnetwork parameters jointly}
    \State $\boldsymbol{\Theta} \leftarrow \boldsymbol{\Theta} - \eta_{\mathrm{sched}} \, \nabla_{\boldsymbol{\Theta}} \mathcal{L}_{\mathrm{total}}$
 
\EndFor
\State \Return trained model $\{\hat{u}_i(t)\}_{i=1}^{N}$
\end{algorithmic}
\end{algorithm}

\begin{algorithm}[H]
\caption{Original NA-PINN Applied to the Cascading Tank System~\cite{shin2025napinn}}
\label{alg:napinn_tank}
\begin{algorithmic}[1]
\Require Number of tanks $n$, collocation points $\{t_i\}_{i=1}^{N_t}$ on $[0,\, T]$
\Require Initial conditions: $\{H_{0,i}\}_{i=1}^{n}$ (water levels), $\{v_{0,j}\}_{j=1}^{n-1}$ (velocities)
\Require Subnetwork parameters $\boldsymbol{\Theta} = \{ \boldsymbol{\theta}_{\hat{H}_1}, \dots, \boldsymbol{\theta}_{\hat{H}_{n}},\, \boldsymbol{\theta}_{\hat{v}_1}, \dots, \boldsymbol{\theta}_{\hat{v}_{n-1}} \}$
\Require Number of epochs $N_{\mathrm{epoch}}$, learning rate scheduler $\eta_{\mathrm{sched}}$, loss weights $\lambda_1,\, \lambda_2$
 
\State Initialize all subnetwork parameters $\boldsymbol{\Theta}$
 
\Statex \hspace{\algorithmicindent} \textit{// Enforce initial conditions via shifting (hard constraint)}
\State $\hat{H}_i(t) = \mathrm{NN}_{\boldsymbol{\theta}_{\hat{H}_i}}(t) - \mathrm{NN}_{\boldsymbol{\theta}_{\hat{H}_i}}(0) + H_{0,i}$ \quad for $i = 1, \dots, n$
\State $\hat{v}_j(t) = \mathrm{NN}_{\boldsymbol{\theta}_{\hat{v}_j}}(t) - \mathrm{NN}_{\boldsymbol{\theta}_{\hat{v}_j}}(0) + v_{0,j}$ \quad for $j = 1, \dots, n{-}1$
 
\For{epoch $= 1, \dots, N_{\mathrm{epoch}}$}
 
    \State Predict $\{ \hat{H}_{i}(t') \}_{i=1}^{n}$ and $\{ \hat{v}_{j}(t') \}_{j=1}^{n-1}$ using each subnetwork
 
    \Statex \hspace{\algorithmicindent}\hspace{\algorithmicindent} \textit{// Compute temporal derivatives via automatic differentiation}
    \State $\dfrac{\partial \hat{v}_{j}}{\partial t}$ for $j = 1, \dots, n{-}1$, \quad $\dfrac{\partial \hat{H}_{i}}{\partial t}$ for $i = 1, \dots, n$
 
    \Statex \hspace{\algorithmicindent}\hspace{\algorithmicindent} \textit{// Evaluate momentum residuals for each flow path $j$}
    \State $\mathcal{R}_{\mathrm{momentum},\,j} = L \dfrac{\partial \hat{v}_{j}}{\partial t} - g \, \Delta z_j + \dfrac{K^*}{2} |\hat{v}_{j}|\, \hat{v}_{j}$
 
    \Statex \hspace{\algorithmicindent}\hspace{\algorithmicindent} \textit{// Evaluate continuity residuals for each tank $i$}
    \State $\mathcal{R}_{\mathrm{continuity},\,i} = \rho A_t \dfrac{\partial \hat{H}_{i}}{\partial t} +
    \begin{cases}
    \rho A_p F \, \hat{v}_{1} (1 - \alpha_1), & i = 1 \\
    \rho A_p F \left[ \hat{v}_{i} (1 - \alpha_i) - \hat{v}_{i-1} (1 - \alpha_{i-1}) \right], & 2 \le i \le n{-}1 \\
    -\rho A_p F \, \hat{v}_{n-1} (1 - \alpha_{n-1}), & i = n
    \end{cases}$
 
    \Statex \hspace{\algorithmicindent}\hspace{\algorithmicindent} \textit{// Compute total loss}
    \State $\mathcal{L}_{\mathrm{PDE}} = \lambda_1 \cdot \dfrac{1}{N_t} \displaystyle\sum_{j=1}^{n-1} \mathcal{R}_{\mathrm{momentum},\,j}^{\,2} + \lambda_2 \cdot \dfrac{1}{N_t} \displaystyle\sum_{i=1}^{n} \mathcal{R}_{\mathrm{continuity},\,i}^{\,2}$
 
    \Statex \hspace{\algorithmicindent}\hspace{\algorithmicindent} \textit{// Update all subnetwork parameters jointly}
    \State $\boldsymbol{\Theta} \leftarrow \boldsymbol{\Theta} - \eta_{\mathrm{sched}} \, \nabla_{\boldsymbol{\Theta}} \mathcal{L}_{\mathrm{PDE}}$
 
\EndFor
\State \Return trained model $\{\hat{H}_i(t),\, \hat{v}_j(t)\}$
\end{algorithmic}
\end{algorithm}
 
\noindent Under comparable total parameter counts, NA-PINN reduced the water height MSE by over five orders of magnitude relative to the shared-network baseline across all tested configurations ($N_i = 2$, 3, and 6).
 
These results established the feasibility of a data-free PINN approach for MELCOR's thermal-hydraulic modules.
However, NA-PINN exhibits a practical limitation that motivates the present study:
\begin{itemize}
    \item \textbf{Fixed-scenario limitation:} A trained NA-PINN is valid only for one specific set of initial conditions. Changing any parameter---such as the initial water level in a CV---requires complete retraining, preventing its use as a surrogate model.
\end{itemize}
More broadly, the long-horizon limitation discussed in Section~\ref{sec:1} further motivates a hybrid coupling strategy, detailed in Section~\ref{sec:2.4}.
The present study pursues these goals through two complementary strategies: parameterization of the PINN input space (Section~\ref{sec:2.3}) and hybrid coupling with an FDM solver (Section~\ref{sec:2.4}).

\subsection{Parameterized PINN}\label{sec:2.3}

The fixed-scenario limitation identified above is not unique to NA-PINN but is inherent to standard PINN formulations, which encode a single set of conditions into the trained parameters.
Parameterized PINNs address this by extending the network input space to include physical parameters $\boldsymbol{\mu}$ alongside the spatiotemporal coordinates, so that the network learns a solution manifold $u(\mathbf{x}, t, \boldsymbol{\mu}; \boldsymbol{\theta})$ spanning a prescribed parameter range.
Once trained, such a network can predict solutions for any parameter combination within the training range without retraining.
 
Cho et al.\ \cite{cho2024} formalized this concept as Parameterized Physics-Informed Neural Networks (P$^2$INNs), employing separate encoder subnetworks for coordinates and parameters whose outputs are fused through a decoder.
While this dual-encoder architecture offers strong representational capacity, a simpler alternative is to directly concatenate the parameters with the spatiotemporal inputs of a standard fully-connected network, avoiding the need for separate encoder subnetworks.
Through extensive evaluation on benchmark 1D and 2D parameterized PDEs, P$^2$INNs outperformed conventional PINN baselines in both accuracy and parameter efficiency while mitigating known failure modes.
Training is performed by sampling collocation points jointly in the $(\mathbf{x}, t, \boldsymbol{\mu})$ space, enabling the network to learn how the solution varies continuously across the parameter domain.
Interpolation within the trained range is generally accurate, whereas extrapolation beyond the training boundaries remains limited.
This constraint implies that the training parameter range must be chosen to cover the expected operating conditions; the present study addresses this by defining a physically motivated sampling range for the initial conditions of interest (Section~\ref{sec:3}).
 
Within the nuclear engineering domain, Xie et al.\ \cite{xie2024} applied parameterized PINNs to the steady-state mono-energy neutron diffusion equation, achieving acceleration ratios exceeding $1000\times$ compared to conventional solvers under hard boundary constraints.
In the broader computational physics community, Donnelly et al.\ \cite{donnelly2024} developed a PINN-based surrogate for hydrodynamic simulators governed by the shallow water equations, parameterizing boundary conditions as network inputs and outperforming purely data-driven approaches by up to 25\%.
Arthurs and King \cite{arthurs2021} demonstrated the interpolation of parametric Navier--Stokes solutions using actively trained PINNs.
 
These studies collectively show that parameterized PINNs can serve as effective surrogates for various PDE systems.
Among them, Cho et al.\ \cite{cho2024} and Xie et al.\ \cite{xie2024} demonstrated fully data-free training, while Donnelly et al.\ \cite{donnelly2024} and Arthurs and King \cite{arthurs2021} incorporated simulation data with physics-informed regularization.
However, none of these approaches---whether data-free or data-assisted---have been applied to lumped-parameter thermal-hydraulic system codes such as MELCOR.

\subsection{Research Gap}\label{sec:2.4}

The literature reviewed in the preceding subsections reveals two distinct gaps in the current state of the art.
 
\textbf{Gap 1: No data-free surrogate for nuclear system codes.}
Data-driven surrogates for MELCOR and similar codes \cite{chae2023, wangma2025, wangma2023, songha2022, lee2024surrogate, song2025dose} offer parametric flexibility and rapid inference, but they depend on large volumes of simulation data and lack explicit physics-based constraints.
Even the physics-informed surrogate of Baraldi et al.\ \cite{baraldi2025}, which incorporates physical constraints through allocation points, still requires RELAP5-3D simulation data for training.
On a larger scale, the EU Horizon ASSAS project \cite{poubeau2024assas} is actively developing ML-based surrogate models to replace selected modules of the ASTEC and MELCOR severe accident codes; however, these surrogates are also trained on code-generated databases and thus remain data-driven.
Conversely, data-free PINNs---including NA-PINN (Section~\ref{sec:2.2})---cannot function as surrogates due to their fixed-scenario limitation.
Meanwhile, parameterized PINNs have demonstrated surrogate capability in other domains \cite{cho2024, xie2024, donnelly2024}, yet have not been applied to lumped-parameter system codes.
To the best of the authors' knowledge, no existing framework combines data-free training (no simulation data) with surrogate-level parametric flexibility (no retraining) for thermal-hydraulic system code modeling.
 
\textbf{Gap 2: No data-free, node-wise hybrid PINN--numerical coupling.}
Hybrid PINN--numerical frameworks have been explored along two lines: (i) replacing the differential operator computation within PINNs, e.g., substituting automatic differentiation with finite-difference stencils \cite{hfdpinn2025}, and (ii) spatial domain decomposition, either among multiple PINNs~\cite{shukla2021} or between a PINN and a conventional numerical solver~\cite{beitalmal2025}.
Most recently, Wang et al.\ \cite{wang2025tmno} proposed a time-marching framework that couples a physics-informed DeepONet with FEM via the Schwarz alternating method, assigning computationally intensive spatial subdomains to the neural operator while FEM handles the remainder.
On the other hand, the ASSAS project \cite{poubeau2024assas} explores a module-wise hybrid strategy in which data-driven surrogates replace selected physical modules of ASTEC while the remaining modules continue to run as conventional solvers, with data exchanged at each macro time step.
 
However, as discussed in Section~\ref{sec:1}, standalone PINNs and PINN-based surrogates are known to suffer from error accumulation in long-horizon time integration~\cite{wang2024respecting, chen2024pinn}---a limitation that is particularly critical for severe accident simulations spanning hours to days of physical time.
None of the hybrid frameworks reviewed above is specifically designed to address this long-horizon convergence problem.
The present work proposes a hybrid strategy motivated directly by this challenge: a PINN surrogate is assigned to the FP nodes and coupled with an FDM solver at the CV nodes within a shared time-marching loop.
Because the PINN operates within the FDM time-stepping framework, it only needs to predict over a single short time step rather than integrating over the entire temporal domain, thereby inherently eliminating the source of error accumulation.
No study has yet realized such a coupling using a \textit{data-free} PINN surrogate alternating with a conventional numerical solver within a time-marching loop.
 
The present study addresses both gaps simultaneously through the P2F method, combining parameterization with node-assigned hybrid coupling.
A parameterized PINN, extending the NA-PINN design philosophy, is developed to serve as a data-free surrogate at the FP nodes in MELCOR's CVH/FP package (addressing Gap~1).
This network is trained once and applied across all FP nodes, then coupled with an FDM solver that advances the solution at the CV nodes at each time step (addressing Gap~2).
The resulting framework requires no simulation data, generalizes across initial conditions without retraining, and is directly compatible with the node-based structure of existing system codes.

\section{Parameterized NA-PINN for MELCOR CVH/FP Module}\label{sec:3}

Unlike in our previous work~\cite{shin2025napinn}, where a dedicated sub-network was assigned to each output variable, the present study assigns the entire network to a single physical node.
An identically trained NA-PINN is then deployed at each FP node within the CVH/FP module and coupled with the FDM solver within a time-marching loop.
Specifically, the parameterized NA-PINN takes the water-level difference $\Delta h$ between two adjacent CVs and the initial velocity $v_o$ at $t=0$ as inputs, and a single trained model serves as the momentum conservation equation solver for computing the velocity $v$ across all FLs.
This design preserves the core principle of NA-PINN---dedicating a network to each physical node---while extending its role from a fixed-scenario solver to a reusable surrogate.

\subsection{MELCOR CVH/FP Package: Governing Equations}\label{sec:3.1}

\begin{figure}[H]
    \centering
    \includegraphics[width=0.85\linewidth]{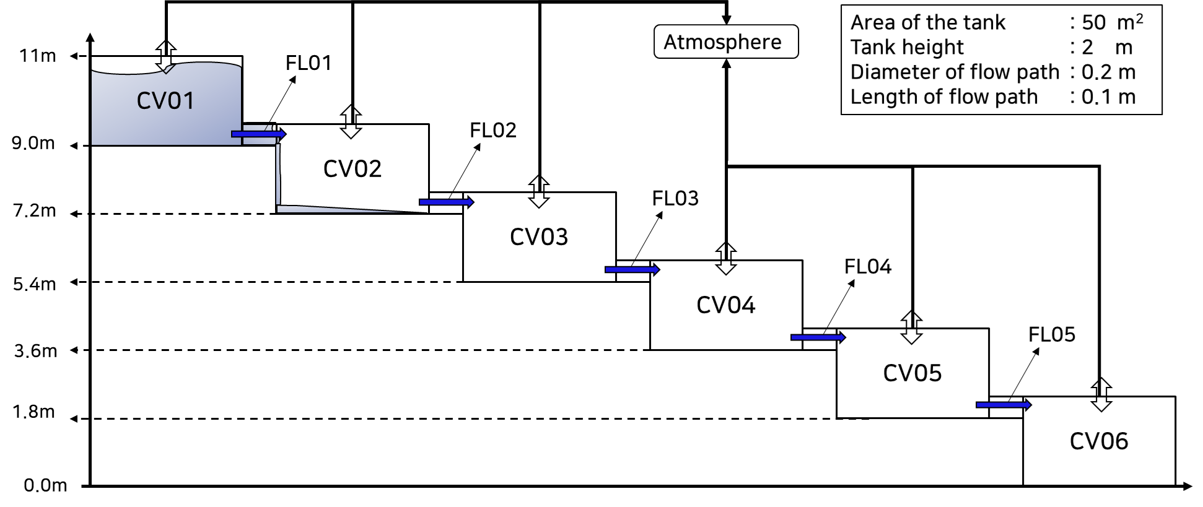}
    \caption{Nodalization for the scenario model.}
    \label{fig:nodalization}
\end{figure}

The thermal-hydraulic module of MELCOR resolves fluid transport across interconnected CVs and FLs by coupling mass and momentum conservation equations. The scenario model considered in this work consists of six cascading open tanks (CV01--CV06) connected by five flow paths (FL01--FL05), as illustrated in Figure~\ref{fig:nodalization}. Each tank has a cross-sectional area of 50\,m$^2$ and a height of 2\,m, while each connecting flow path has a diameter of 0.2\,m and a length of 0.1\,m. The tanks are arranged in a staircase configuration with an elevation drop of 1.8\,m between successive CVs, and all CVs are open to the atmosphere. Water drains from the uppermost tank through the series of FLs to the lowermost tank under gravity.
 
The mass conservation equation for material $m$ within CV $i$ is expressed as:

\begin{align}
\frac{\partial M_{i,m}}{\partial t} &= 
\sum_{j} \sigma_{ij} \alpha_{j,\phi} \rho_{j,m}^d v_{j,\phi} F_j A_j + \dot{M}_{i,m}
\label{eq:mass_pde}
\end{align}

\noindent Here, the index $j$ runs over all FLs connected to CV $i$. The coefficient $\sigma_{ij}$ indicates the direction of flow, $\alpha_{j,\phi}$ is the volume fraction of phase $\phi$, $\rho_{j,m}^d$ is the donor density of material $m$, $v_{j,\phi}$ is the flow velocity, $F_j$ is the fraction of area opened, and $A_j$ is the FP area. The source term $\dot{M}_{i,m}$ represents non-flow mass sources including condensation, evaporation, and fog precipitation. In the present scenario, the cross-sectional areas and the density of water are assumed constant; hence, the mass in each CV can be represented in terms of the water level.

The momentum equation for phase $\phi$ in FP $j$ takes the following partial differential form:

\begin{align}
\alpha_{j,\phi} \rho_{j,\phi} L_j \frac{\partial v_{j,\phi}}{\partial t} &=
\alpha_{j,\phi} (P_i - P_k) +
\alpha_{j,\phi} (\rho g \Delta z)_{j,\phi} +
\alpha_{j,\phi} \Delta P_j +
\alpha_{j,\phi} \rho_{j,\phi} v_{j,\phi} (\Delta v_{j,\phi})  \notag \\
&\quad- \frac{1}{2} K_{j,\phi}^* \alpha_{j,\phi} \rho_{j,\phi} |v_{j,\phi}| v_{j,\phi} 
- \alpha_{j,\phi} \alpha_{j,-\phi} f_{2,j} L_{2,j} (v_{j,\phi} - v_{j,-\phi})
\label{eq:momentum_pde}
\end{align}

\noindent In this expression, subscripts $i$ and $k$ denote the donor CV and the receiver CV, respectively, $L_j$ is the inertial length, $\Delta P_j$ is the pump head pressure, $K_{j,\phi}^*$ is the net form- and wall-loss coefficient, and $f_{2,j}$ and $L_{2,j}$ are the momentum exchange coefficient and the effective length over which the interphase force acts. In conventional FDM-based system codes such as MELCOR, Eq.~\eqref{eq:momentum_pde} is discretized and solved iteratively at each time step. By contrast, the NA-PINN directly employs the continuous PDE form of the momentum equation as its physics-informed loss function, where the temporal derivative $\partial v / \partial t$ is computed via automatic differentiation rather than finite-difference approximation.

In the original MELCOR formulation, the momentum equations for all FLs are coupled through the shared CV pressures and water levels, forming a system of simultaneous algebraic equations that is assembled into a matrix and solved globally at each time step.
This coupling arises because the velocity update at one FP depends on the pressure and level changes induced by flows through adjacent FLs.
In the present work, however, the simplified momentum equation---obtained under the open-tank assumption where the inter-CV pressure differential is negligible---decouples the individual FP equations: the velocity at each FP depends only on the local water-level difference $\Delta h$ and the local friction term, with no implicit coupling to other FLs.
Consequently, the momentum equation can be evaluated sequentially
for each flow path rather than simultaneously, and this sequential structure is precisely what enables the deployment of a single parameterized PINN that processes each FP independently with its own local inputs $(\Delta h_j,\, v_{o,j},\, t)$. In the proposed framework, only the momentum conservation equation is employed in its continuous form for training the NA-PINN, while the mass conservation equation is evaluated in its discretized form by the emulator during the simulation phase.

\subsection{Parameterized NA-PINN Architecture for the CVH/FP Module}\label{sec:3.2}

As one of the most effective approaches to addressing the coupling challenge between strongly coupled modules in PINN--numerical hybrid methodologies, the present study proposes a parameterized NA-PINN architecture. To enable node-assigned coupling with FDM within a time-marching loop, the NA-PINN accepts $\Delta h$ and $v_o$ as parametric inputs, thereby satisfying the conditions of a surrogate model, and a single trained network is reused across each flow path within the CVH/FP module, predicting the velocity of each FP individually. The training time domain $T$ is set to match the maximum allowable time step $\Delta t_{\max}$ of the FDM solver, so that each PINN inference directly produces the velocity update for one time step of the hybrid simulation.

\begin{figure}[H]
    \centering
    \includegraphics[width=0.85\linewidth]{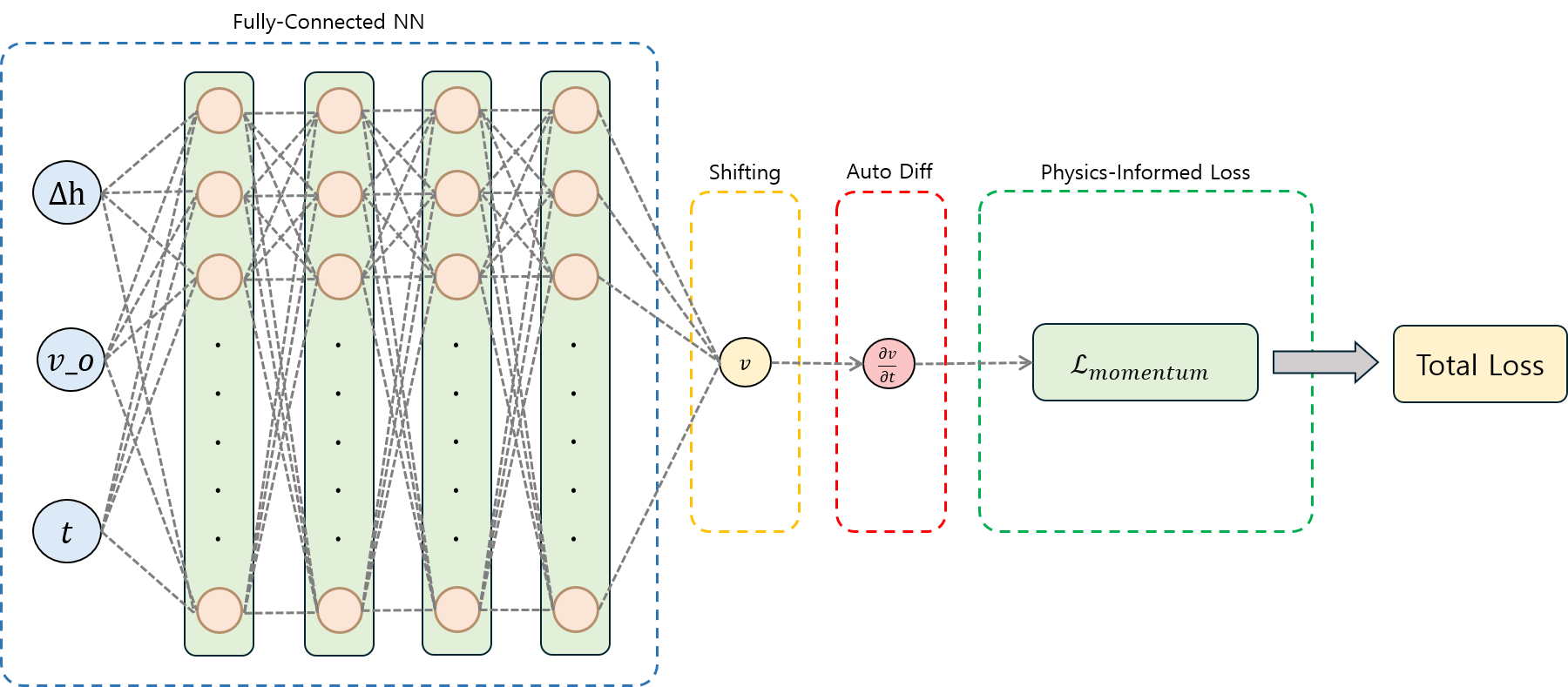}
    \caption{Architecture of the reformulated NA-PINN for a single FP node. The FCN takes $\Delta h$, $v_o$, and $t$ as inputs and predicts the velocity $v$, from which $\partial v / \partial t$ is computed via automatic differentiation. The momentum residual $\mathcal{L}_{\mathrm{momentum}}$ is evaluated and accumulated into the total loss.}
    \label{fig:NA-PINN_architecture}
\end{figure}

All three inputs are normalized prior to the forward pass to ensure that the input space spans a comparable numerical range: $\bar{h} = \Delta h / \Delta h_{\mathrm{train}}$, $\bar{t} = t / T$, and $\bar{v}_0 = v_0 / v_{0,\max}$, where $\Delta h_{\mathrm{train}}$, $T$, and $v_{0,\max}$ denote the respective training-domain bounds. This normalization prevents any single input dimension from dominating the gradient landscape during training.

A common challenge in PINN training is the enforcement of initial and boundary conditions. The conventional soft-constraint approach appends additional penalty terms to the loss function, which introduces competing objectives and requires careful tuning of the associated weighting coefficients. To avoid this difficulty, we adopt a hard constraint formulation that embeds the initial condition directly into the network output:

\begin{equation}
    \hat{v}(t) = v_0 + t \cdot \mathrm{NN}_{\boldsymbol{\theta}}(\bar{h},\, \bar{t},\, \bar{v}_0)
\label{eq:hard_ic}
\end{equation}

\noindent By construction, Eq.~\eqref{eq:hard_ic} satisfies $\hat{v}(0) = v_0$ exactly for any set of network parameters $\boldsymbol{\theta}$. This eliminates the need for an initial-condition loss term entirely, allowing the optimizer to focus exclusively on minimizing the physics-based residual. Furthermore, the multiplicative coupling with $t$ ensures that the network output contributes progressively as time advances, which stabilizes the early stages of training where the residual gradients would otherwise be dominated by the initial-condition penalty.

As noted in Section~\ref{sec:3.1}, the NA-PINN directly employs the continuous PDE form of the momentum conservation equation as its loss function, rather than a discretized approximation. The training is conducted in a purely data-free manner. For each collocation point $i$, the momentum residual is evaluated as:

\begin{equation}
    \mathcal{R}_{\mathrm{momentum},\,i} = L \frac{\partial \hat{v}_i}{\partial t} - g \, \Delta h_i + \frac{K^*}{2} |\hat{v}_i|\, \hat{v}_i
\label{eq:momentum_residual}
\end{equation}

\noindent where $L$ is the inertial length of the flow path, $g$ is gravitational acceleration, $\Delta h_i$ is the water-level difference, and $K^*$ is the net form- and wall-loss coefficient. The first term represents the temporal acceleration, the second term captures the gravitational driving force through the static head difference, and the third term accounts for the hydraulic friction loss. The total training loss is then the mean squared residual over $N_b$ collocation points:

\begin{equation}
    \mathcal{L}_{\mathrm{total}} = \frac{1}{N_b} \sum_{i=1}^{N_b} \mathcal{R}_{\mathrm{momentum},\,i}^{\,2}
\label{eq:total_loss}
\end{equation}

\noindent Because the hard initial condition (Eq.~\eqref{eq:hard_ic}) already enforces $\hat{v}(0) = v_0$, the loss function in Eq.~\eqref{eq:total_loss} consists of a single term with no auxiliary penalty, avoiding the multi-objective balancing issue that commonly arises in PINN training.

The collocation points are generated once before training and held fixed throughout the entire optimization process, rather than being resampled at each epoch. For a batch of $N_b$ points, the water-level difference $h_i$ is drawn from $\mathcal{U}[0,\, \Delta h_{\mathrm{train}}]$ for a fraction $(1 - r_{h_0})$ of the batch, while the remaining $r_{h_0}$ fraction is fixed at $h_i = 0$ to ensure that the network adequately learns the zero-flow regime where no driving head exists. Similarly, the initial velocity $v_{0,i}$ is sampled from $\mathcal{U}[0,\, v_{0,\max}]$ for a fraction $(1 - r_{v_0})$, with the remaining $r_{v_0}$ fraction set to $v_{0,i} = 0$ to reinforce the quiescent initial state. The time coordinate $t_i$ is uniformly sampled from $\mathcal{U}[0,\, T]$ for all points. The rationale for fixing the collocation set is twofold. First, a fixed set enables consistent evaluation of the validation loss across epochs, which is used to track training progress and to restore the best-performing model. Second, it eliminates the stochastic noise introduced by resampling, leading to smoother convergence trajectories. The boundary-enriched sampling with ratios $r_{h_0}$ and $r_{v_0}$ serves as a form of importance sampling, concentrating a controlled fraction of training effort on the physically critical boundaries where the velocity must vanish.

The network parameters are optimized using a first-order gradient-based optimizer with a piecewise learning rate schedule that reduces the step size at predefined epoch milestones to facilitate fine-grained convergence in the later stages of training. Gradient clipping is applied at each update step to prevent instability arising from large residual gradients, particularly during early epochs when the network output may deviate substantially from the physical solution. A separate validation collocation set, sampled with the same strategy as the training set, is evaluated periodically, and the model state corresponding to the lowest validation loss is retained. The complete training procedure is summarized in Algorithm~\ref{alg:reformulated_napinn}.

\begin{algorithm}[H]
\caption{Training Procedure of the Reformulated (Parameterized) NA-PINN}
\label{alg:reformulated_napinn}
\begin{algorithmic}[1]
\Require PDE $\mathcal{N}[u] = 0$ governing FP node dynamics
\Require Parameterized input domain: IC parameter $p \in [0,\, p_{\max}]$, BC parameter $u_0 \in [0,\, u_{0,\max}]$, $t \in [0,\, T]$
\Require Single FCN parameters $\boldsymbol{\theta}$ (shared across all FP nodes)
\Require Collocation points $\{(p_i,\, t_i,\, u_{0,i})\}_{i=1}^{N_b}$, number of epochs $N_{\mathrm{epoch}}$
\Require Learning rate schedule $\{(e_k,\, \eta_k)\}_{k=1}^{K}$
 
\State Initialize FCN parameters $\boldsymbol{\theta}$
 
\For{epoch $= 1, \dots, N_{\mathrm{epoch}}$}
 
    \Statex \hspace{\algorithmicindent}\hspace{\algorithmicindent} \textit{// Forward pass: parameterized input $\rightarrow$ single nodal output}
    \State $\hat{u}_i = u_{0,i} + t_i \cdot \mathrm{NN}_{\boldsymbol{\theta}}(\bar{p}_i,\, \bar{t}_i,\, \bar{u}_{0,i})$ \hfill \textit{// hard IC enforcement}
 
    \Statex \hspace{\algorithmicindent}\hspace{\algorithmicindent} \textit{// Compute temporal derivative via automatic differentiation}
    \State $\dfrac{\partial \hat{u}_i}{\partial t}$
 
    \Statex \hspace{\algorithmicindent}\hspace{\algorithmicindent} \textit{// Evaluate PDE residual}
    \State $\mathcal{R}_i = \mathcal{N}[\hat{u}_i,\, p_i]$
 
    \Statex \hspace{\algorithmicindent}\hspace{\algorithmicindent} \textit{// Compute total loss}
    \State $\mathcal{L}_{\mathrm{total}} = \dfrac{1}{N_b} \displaystyle\sum_{i=1}^{N_b} \mathcal{R}_i^{\,2}$
 
    \Statex \hspace{\algorithmicindent}\hspace{\algorithmicindent} \textit{// Update parameters}
    \State $\boldsymbol{\theta} \leftarrow \boldsymbol{\theta} - \eta_k \, \nabla_{\boldsymbol{\theta}} \mathcal{L}_{\mathrm{total}}$
 
\EndFor
\State \Return trained model $\mathrm{NN}_{\boldsymbol{\theta}^*}$ \quad \textit{// applicable to any FP node with arbitrary $(p,\, u_0)$}
\end{algorithmic}
\end{algorithm}

\begin{algorithm}[H]
\caption{Reformulated NA-PINN Applied to the Cascading Tank System}
\label{alg:reformulated_napinn_tank}
\begin{algorithmic}[1]
\Require Training domain $\Delta h \in [0,\, \Delta h_{\mathrm{train}}]$, $v_0 \in [0,\, v_{0,\max}]$, $t \in [0,\, T]$ where $T = \Delta t_{\max}$ of the FDM
\Require Batch size $N_b$, boundary-point ratios $r_{h_0}$ and $r_{v_0}$, number of epochs $N_{\mathrm{epoch}}$
\Require Physical constants $L,\, g,\, K^*$
 
\State Initialize fully connecred network parameters $\boldsymbol{\theta}$
 
\Statex \hspace{\algorithmicindent} \textit{// Generate fixed collocation sets with boundary enrichment (one-time)}
\State Sample $\Delta h_i$: \; $N_b(1{-}r_{h_0})$ from $\mathcal{U}[0,\, \Delta h_{\mathrm{train}}]$, \; remaining $N_b r_{h_0}$ fixed at $0$
\State Sample $t_i$: \; all $N_b$ from $\mathcal{U}[0,\, T]$
\State Sample $v_{0,i}$: \; $N_b(1{-}r_{v_0})$ from $\mathcal{U}[0,\, v_{0,\max}]$, \; remaining $N_b r_{v_0}$ fixed at $0$
\State $\mathcal{C}_{\mathrm{train}} \leftarrow \{(\Delta h_i,\, t_i,\, v_{0,i})\}_{i=1}^{N_b}$
 
\For{epoch $= 1, \dots, N_{\mathrm{epoch}}$}
 
    \Statex \hspace{\algorithmicindent}\hspace{\algorithmicindent} \textit{// Forward pass: $(\Delta h,\, t,\, v_0) \rightarrow$ velocity $v$}
    \State Retrieve $(\Delta h_i,\, t_i,\, v_{0,i})$ from $\mathcal{C}_{\mathrm{train}}$
    \State Normalize: $\bar{h}_i = \Delta h_i / \Delta h_{\mathrm{train}}$
    \State Hard IC: $\hat{v}_i = v_{0,i} + t_i \cdot \mathrm{NN}_{\boldsymbol{\theta}}(\bar{h}_i,\, t_i / T,\, v_{0,i} / v_{0,\max})$
 
    \Statex \hspace{\algorithmicindent}\hspace{\algorithmicindent} \textit{// Compute temporal derivative via automatic differentiation}
    \State $\dfrac{\partial \hat{v}_i}{\partial t} \leftarrow \mathrm{autograd}(\hat{v}_i,\, t_i)$
 
    \Statex \hspace{\algorithmicindent}\hspace{\algorithmicindent} \textit{// Evaluate momentum residual}
    \State $\mathcal{R}_i = L \dfrac{\partial \hat{v}_i}{\partial t} - g \, \Delta h_i + \dfrac{K^*}{2} |\hat{v}_i|\, \hat{v}_i$
 
    \Statex \hspace{\algorithmicindent}\hspace{\algorithmicindent} \textit{// Compute total loss and update parameters}
    \State $\displaystyle\mathcal{L} = \frac{1}{N_b} \sum_{i=1}^{N_b} \mathcal{R}_i^{\,2}$
    \State $\boldsymbol{\theta} \leftarrow \boldsymbol{\theta} - \eta \, \nabla_{\boldsymbol{\theta}} \mathcal{L}$
 
\EndFor
\State \Return trained model $\mathrm{NN}_{\boldsymbol{\theta}^*}$ \quad \textit{// applicable to all FLs with arbitrary $(\Delta h,\, v_0)$}
\end{algorithmic}
\end{algorithm}

In Algorithm~\ref{alg:reformulated_napinn}, the parameterized NA-PINN is trained entirely without data by minimizing the momentum equation residual over fixed collocation points.
The piecewise learning rate schedule progressively reduces the step size at predefined epoch milestones, enabling coarse exploration in the early stages followed by fine-grained convergence toward the end. Gradient clipping bounds the update norm to $\gamma$, preventing divergence when the residual gradients are large. The validation loss is periodically evaluated on an independent collocation set $\mathcal{C}_{\mathrm{val}}$, and the model state with the lowest validation loss is retained, effectively implementing an early stopping strategy without prematurely terminating training. The hard initial condition (Eq.~\eqref{eq:hard_ic}) ensures the exact satisfaction of $\hat{v}(0) = v_0$ without an auxiliary loss term, so that the optimizer concentrates entirely on reducing the physics-based residual. The boundary-enriched collocation strategy (Algorithm~\ref{alg:reformulated_napinn_tank}) allocates a controlled fraction of training points at $h = 0$ and $v_0 = 0$, improving the network's accuracy in the zero-flow regime without degrading its performance in the interior of the input domain. Once training is complete, the resulting model $\mathrm{NN}_{\boldsymbol{\theta}^*}$ can be deployed as a surrogate velocity solver within a time-stepping simulation framework, as described in the following subsection.

\subsection{Standalone Verification of the Parameterized NA-PINN}\label{sec:3.3}

To verify the accuracy of the trained parameterized NA-PINN as a standalone momentum equation solver, its predictions are compared against the reference FDM solution for three representative input conditions that span distinct regions of the training parameter space.
The conditions are selected to cover qualitatively different flow regimes:
$(\Delta h,\, v_o) = (1.0,\, 0)$ represents a moderate driving head with a quiescent initial state; $(\Delta h,\, v_o) = (2.0,\, 3.0)$ combines the maximum driving head with a moderate initial velocity; and $(\Delta h,\, v_o) = (1.0,\, 6.0)$ pairs a moderate driving head with a high initial velocity that substantially exceeds the corresponding equilibrium value. In each case, the PINN receives the specified $(\Delta h,\, v_o)$ as inputs alongside the time coordinate $t \in [0,\,T]$ and outputs the velocity profile $\hat{v}(t)$, which is compared against the FDM solution of the momentum equation~\eqref{eq:momentum_pde} under the same conditions.
 
\begin{figure}[H]
    \centering
    \begin{subfigure}[b]{0.32\textwidth}
        \centering
        \includegraphics[width=\textwidth]{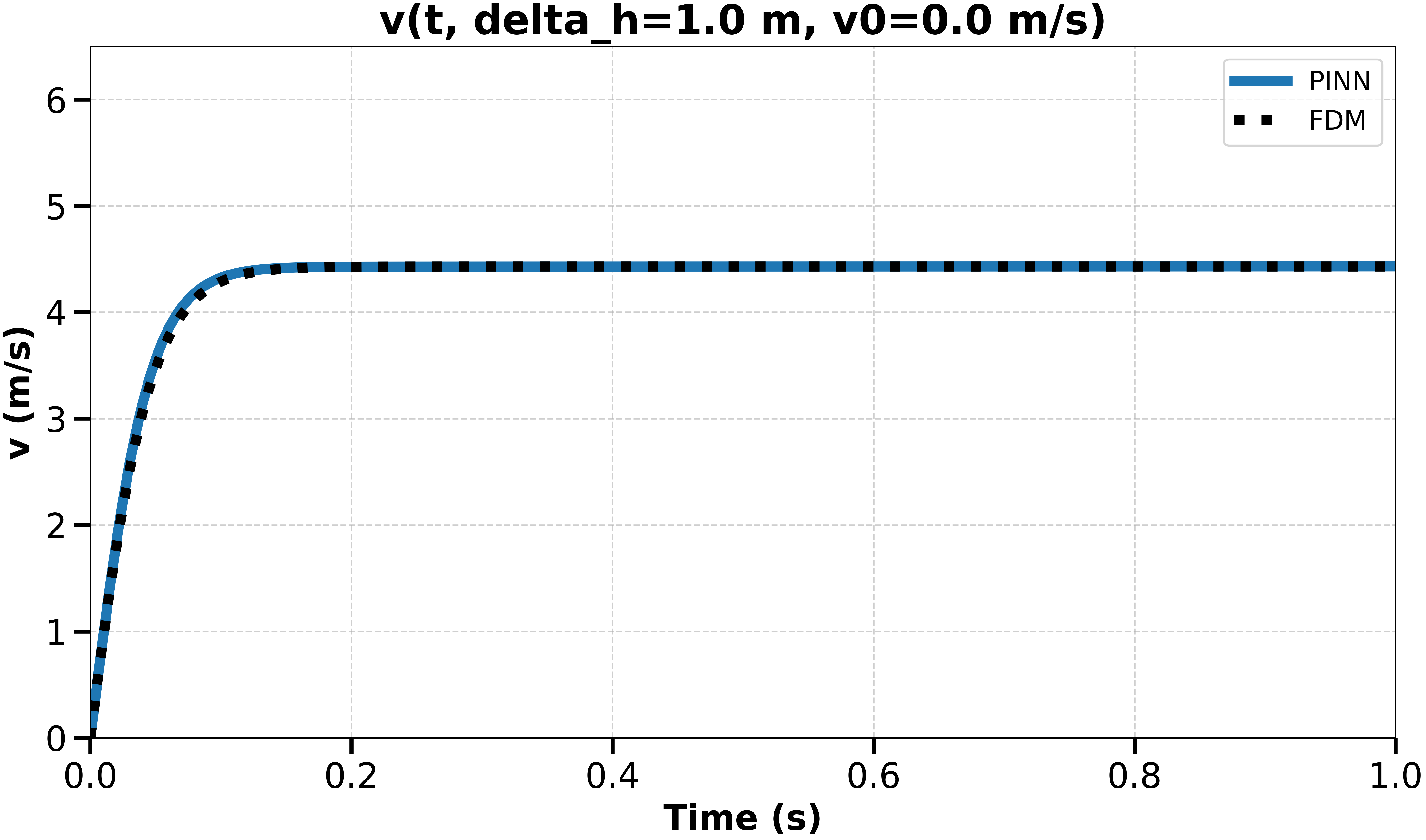}
        \caption{$(\Delta h,\, v_o) = (1.0,\, 0)$}
        \label{fig:pinn_case1}
    \end{subfigure}
    \hfill
    \begin{subfigure}[b]{0.32\textwidth}
        \centering
        \includegraphics[width=\textwidth]{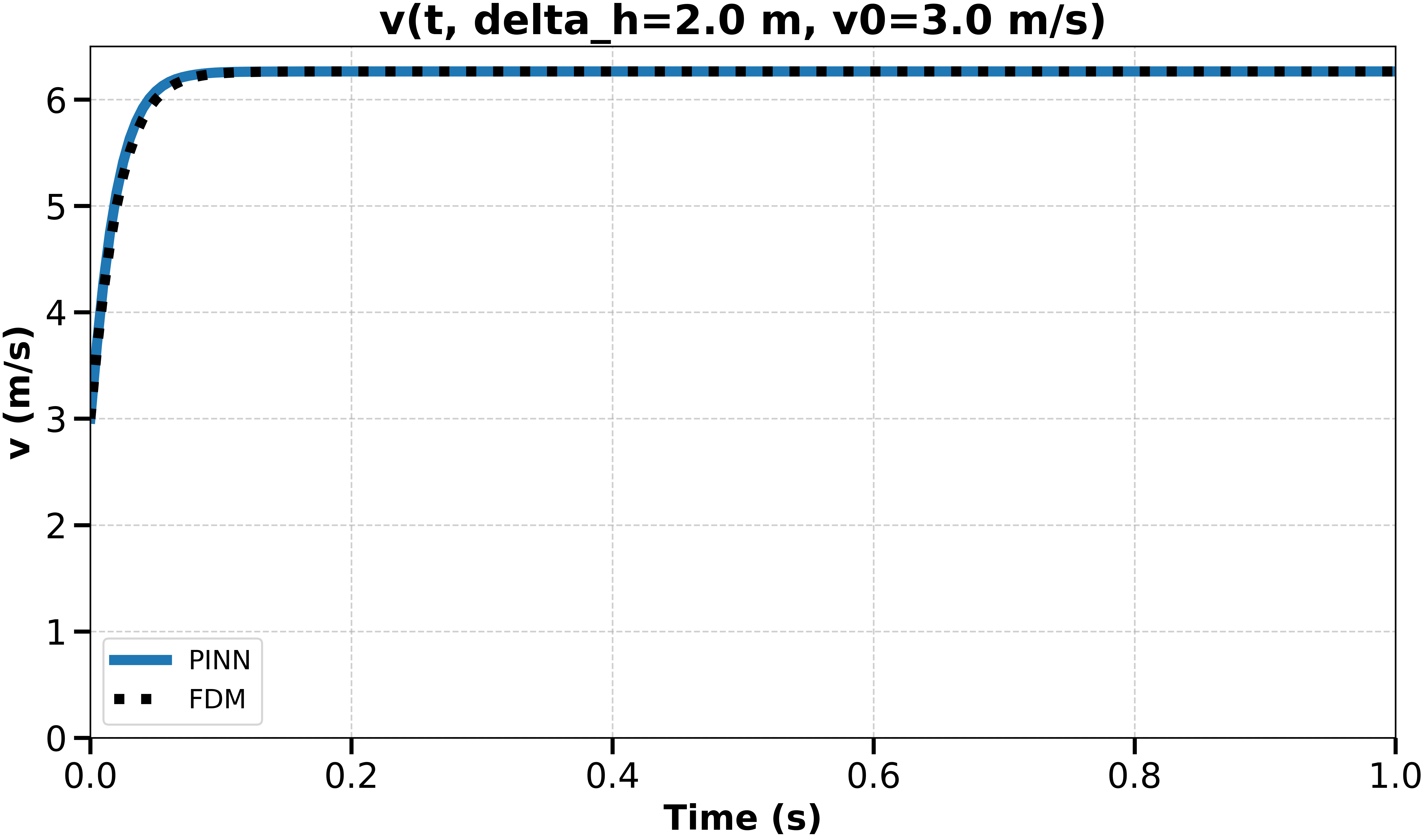}
        \caption{$(\Delta h,\, v_o) = (2.0,\, 3.0)$}
        \label{fig:pinn_case2}
    \end{subfigure}
    \hfill
    \begin{subfigure}[b]{0.32\textwidth}
        \centering
        \includegraphics[width=\textwidth]{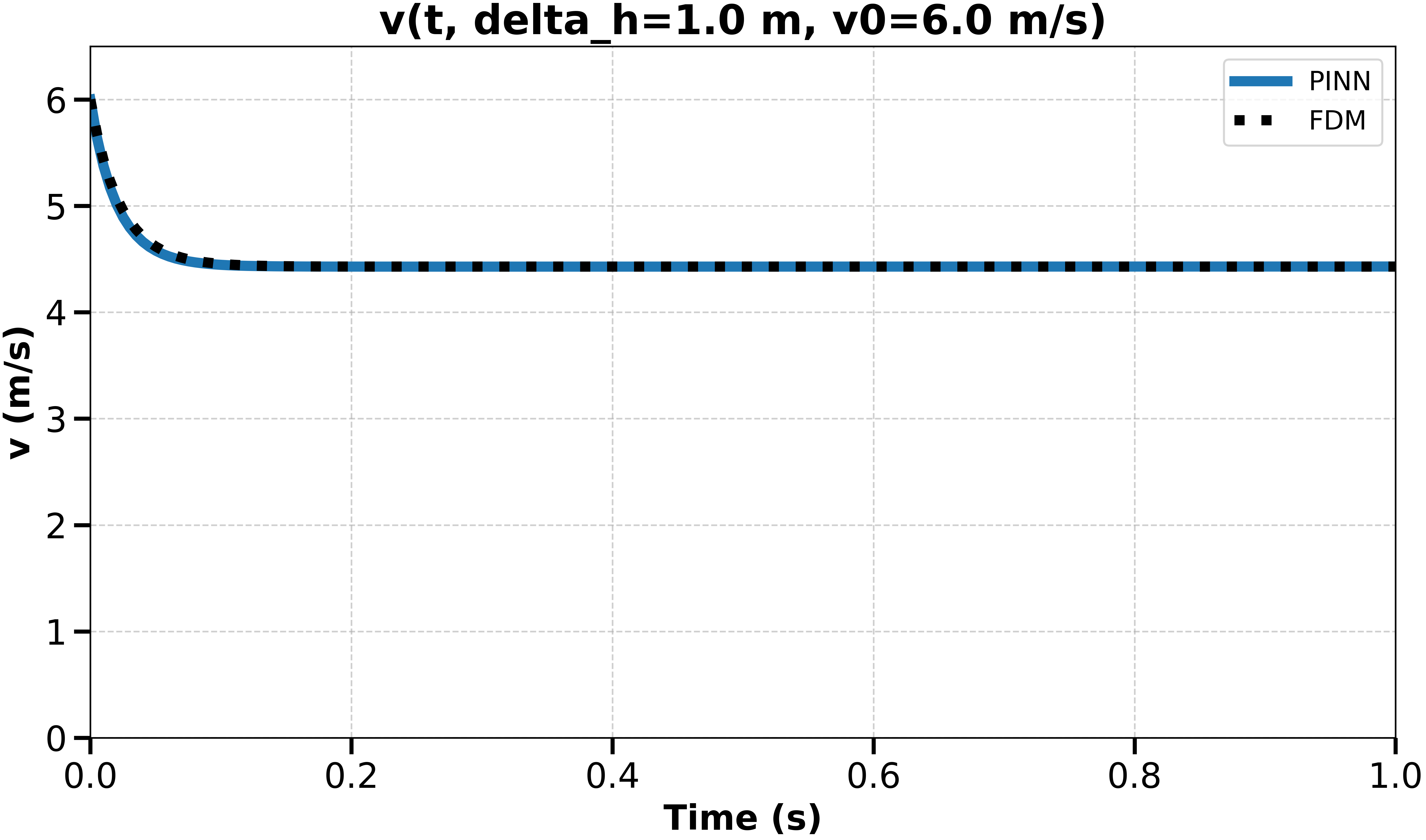}
        \caption{$(\Delta h,\, v_o) = (1.0,\, 6.0)$}
        \label{fig:pinn_case3}
    \end{subfigure}
 
    \caption{Standalone PINN predictions versus reference FDM solutions for three input conditions.
    Solid: parameterized PINN; dashed: reference FDM.}
    \label{fig:pinn_standalone}
\end{figure}

Figure~\ref{fig:pinn_standalone} presents the velocity profiles for the three conditions.
In all three cases, the PINN predictions (solid lines) closely follow the reference FDM solutions (dashed lines) over the entire time window.
For the quiescent case $(\Delta h,\, v_o) = (1.0,\, 0)$ (Fig.~\ref{fig:pinn_case1}), the velocity rises from zero under the gravitational driving force and gradually approaches the steady-state value determined by the balance between the static head and friction loss; the PINN accurately captures both the transient rise and the asymptotic plateau.
For $(\Delta h,\, v_o) = (1.0,\, 6.0)$ (Fig.~\ref{fig:pinn_case3}), the initial velocity far exceeds the equilibrium value corresponding to $\Delta h = 1.0$~m, causing a rapid deceleration as friction dominates over the relatively small driving head; this behavior is likewise captured accurately.
 
Table~\ref{tab:pinn_standalone_errors} quantifies the prediction errors for the three conditions. The MAE and MSE remain consistently low across all three cases, with the MAE of $\mathcal{O}(10^{-3})$~m/s and the MSE of $\mathcal{O}(10^{-4})$~(m/s)$^2$. 
 
\begin{table}[H]
    \centering
    \caption{Prediction errors of the standalone parameterized PINN for three input conditions.}
    \label{tab:pinn_standalone_errors}
    \begin{tabular}{l cc}
        \toprule
        $(\Delta h,\, v_o)$ & MAE (m/s) & MSE (m/s)$^2$ \\
        \midrule
        $(1.0,\; 0)$     & $7.30\times10^{-3}$   & $4.19\times10^{-4}$  \\
        $(2.0,\; 3.0)$   & $6.71\times10^{-3}$   & $5.88\times10^{-4}$  \\
        $(1.0,\; 6.0)$   & $4.10\times10^{-3}$   & $1.70\times10^{-4}$  \\
        \bottomrule
    \end{tabular}
\end{table}
 
These results demonstrate that the parameterized NA-PINN, trained in a purely data-free manner, serves as an accurate surrogate for the momentum equation across the training parameter space, covering qualitatively different flow regimes including quiescent start-up, simultaneous driving head and initial momentum, and over-velocity deceleration. In the following section, this trained network is deployed within an node-assigned hybrid framework, where it provides velocity predictions at each time step while a conventional FDM solver advances the mass conservation equation.

\section{P2F Framework: Node-assigned Hybrid PINN--FDM Coupling}
\label{sec:4}
 
\subsection{Framework Overview}
\label{sec:4.1}
 
\begin{figure}[H]
    \centering
    \includegraphics[width=0.85\linewidth]{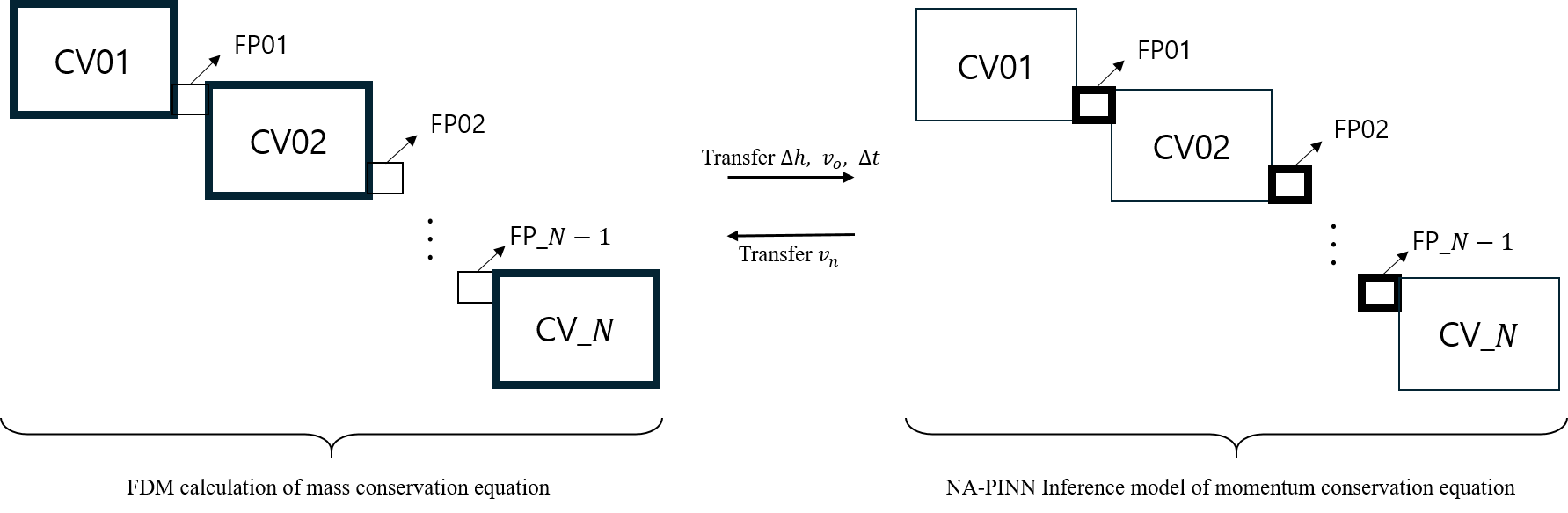}
    \caption{Architecture of the P2F framework. At each time step, the water level in each CV is sequentially updated via the FDM-based mass conservation equation using the predicted velocity $v_n$, and the velocity $v_{n+1}$ in each FP is obtained through the 
    parametrized NA-PINN inference given $\Delta h$, $v_o$, and $\Delta t$.}
    \label{fig:P2F_met_figure}
\end{figure}

The parameterized NA-PINN developed in Section~\ref{sec:3} predicts the FP velocity from the current physical state---the water-level difference $\Delta h$, the previous-step velocity $v_o$, and the elapsed time $t$---for a single flow path.
To advance the full six-tank system over an extended time horizon, these local velocity predictions must be coupled with the mass conservation equation~\eqref{eq:mass_pde} that governs the evolution of CV water levels.
This section introduces the P2F framework(Fig.~\ref{fig:P2F_met_figure})---a node-assigned hybrid strategy that pairs the parameterized PINN with a conventional FDM solver, and verifies the resulting system against reference FDM solutions.

The design principle is to partition the two governing equations according to their numerical character.
The momentum equation~\eqref{eq:momentum_pde}---involving the quadratic 
friction term $|v|\,v$, the implicit static head treatment, and the iterative  friction linearization inherent in the MELCOR FDM formulation---is assigned to the parameterized PINN, replacing the iterative inner loop of the conventional solver with a single forward pass.
The mass conservation equation~\eqref{eq:mass_pde}, which is linear in the water level update once velocities are known, is retained in the FDM solver so that mass is conserved exactly in the discrete sense at every time step.
Because Eq.~\eqref{eq:momentum_pde} is identical in form for every FP, the same trained network serves all five flow paths.
 
At each time step, the procedure consists of four stages:
\begin{enumerate}
    \item \textbf{State evaluation.}
    For each FP $j$, compute the driving head $\Delta h_j^n$ from the current CV water levels $\{h_i^n\}$ and evaluate the void fraction $\alpha_j^n$ at the upstream CV.
    \item \textbf{PINN inference.}
    The parameterized PINN receives $(\Delta h_j^n,\; v_j^n,\; \Delta t)$ and returns the updated velocity $v_j^{n+1}$ via a single forward pass---no iterative solving is required.
    \item \textbf{FDM update.}
    Advance the CV water levels by applying Eq.~\eqref{eq:mass_pde} with the predicted velocities $\{v_j^{n+1}\}$ and void fractions $\{\alpha_j^n\}$.
    \item \textbf{State advance.}
    The updated state $\{h_i^{n+1},\; v_j^{n+1}\}$ replaces the current state, and the loop repeats until the simulation horizon $T$ is reached.
\end{enumerate}

In the original MELCOR formulation, the momentum equations for all flow paths are assembled into a global matrix and solved simultaneously through inner and outer iterations, accounting for bidirectional flow and implicit inter-CV pressure coupling.

In the present study, however, the simplified open-tank scenario ensures a fixed flow direction from upstream to downstream CVs, allowing the coupling algorithm to be constructed in an upwind-scheme fashion where each FP is evaluated sequentially rather than simultaneously.
This sequential structure is what enables the deployment of a single parameterized NA-PINN that processes each FP independently with its own local inputs $(\Delta h_j,\, v_{o,j},\, t)$.

Extension to the full MELCOR formulation---including bidirectional flow, implicit pressure coupling, and the corresponding matrix-based solution strategy---is reserved for future work alongside more complex problem configurations.

Algorithm~\ref{alg:hybrid} formalizes this procedure.
 
\begin{algorithm}[H]
\caption{P2F Time-Marching Procedure}
\label{alg:hybrid}
\begin{algorithmic}[1]
\Require Trained NA-PINN $\mathrm{NN}_{\boldsymbol{\theta}^*}$, number of CVs $n$
\Require Initial water levels $\{h_i^0\}_{i=1}^{n}$, initial velocities $\{v_j^0\}_{j=1}^{n-1}$
\Require Time step $\Delta t$, total simulation time $T_{\mathrm{sim}}$, physical constants $\rho,\, A_t,\, A_p$

\State $N_T \gets \lfloor T_{\mathrm{sim}} / \Delta t \rfloor$

\For{$n = 0$ to $N_T - 1$}

    \Statex \hspace{\algorithmicindent}\hspace{\algorithmicindent} \textit{// PINN inference: velocity prediction for each FL}
    \For{each FL $j = 1, \dots, n-1$ connecting CV $i$ to CV $k$}
        \State Compute $\Delta h_j^n$ and void fraction $\alpha_j^n$ from $\{h_i^n\}$
        \State $v_j^{n+1} \leftarrow \mathrm{NN}_{\boldsymbol{\theta}^*}(\Delta h_j^n / \Delta h_{\mathrm{train}},\; \Delta t,\; v_j^n)$
    \EndFor

    \Statex \hspace{\algorithmicindent}\hspace{\algorithmicindent} \textit{// FDM mass balance: water level update for each CV}
    \For{each CV $i = 1, \dots, n$}
        \State Update $h_i^{n+1}$ via discretized mass conservation (Eq.~\eqref{eq:mass_pde}) using $\{v_j^{n+1}\}$ and $\{\alpha_j^n\}$
    \EndFor

\EndFor
\State \Return $\{h_i^n,\, v_j^n\}$ for $n = 0, 1, \dots, N_T$
\end{algorithmic}
\end{algorithm}
 
An important practical feature is that the inference time step $\Delta t$ need not coincide with the training time window $T$.
During training, the network learns velocity evolution over $[0,\,T]$; at inference, only the single point $t = \Delta t$ is queried, and the updated state is fed back as the initial condition for the next step.
This allows $\Delta t$ to be varied at runtime without retraining, provided $\Delta t \leq T$.

\subsection{Verification and Time-Step Robustness}
\label{sec:4.2}
 
The hybrid framework is verified under a nominal initial condition in which only CV01 is filled ($h_1^0 = 2.0$~m) while all other CVs are empty, with all FP velocities initialized to zero.
This configuration produces the most demanding transient in the six-tank system: the full 2.0~m head across FL01 generates a rapid initial acceleration, followed by a cascading wave of draining and filling that propagates sequentially through all CVs and FLs.
To assess both the coupling accuracy and the time-step flexibility, the nominal case is simulated with three different time steps---$\Delta t = 0.2$, $0.5$, and $1.0$~s---and compared against the reference FDM solution.
 
\begin{figure}[H]
    \centering
    \begin{subfigure}[b]{0.48\textwidth}
        \centering
        \includegraphics[width=\textwidth]{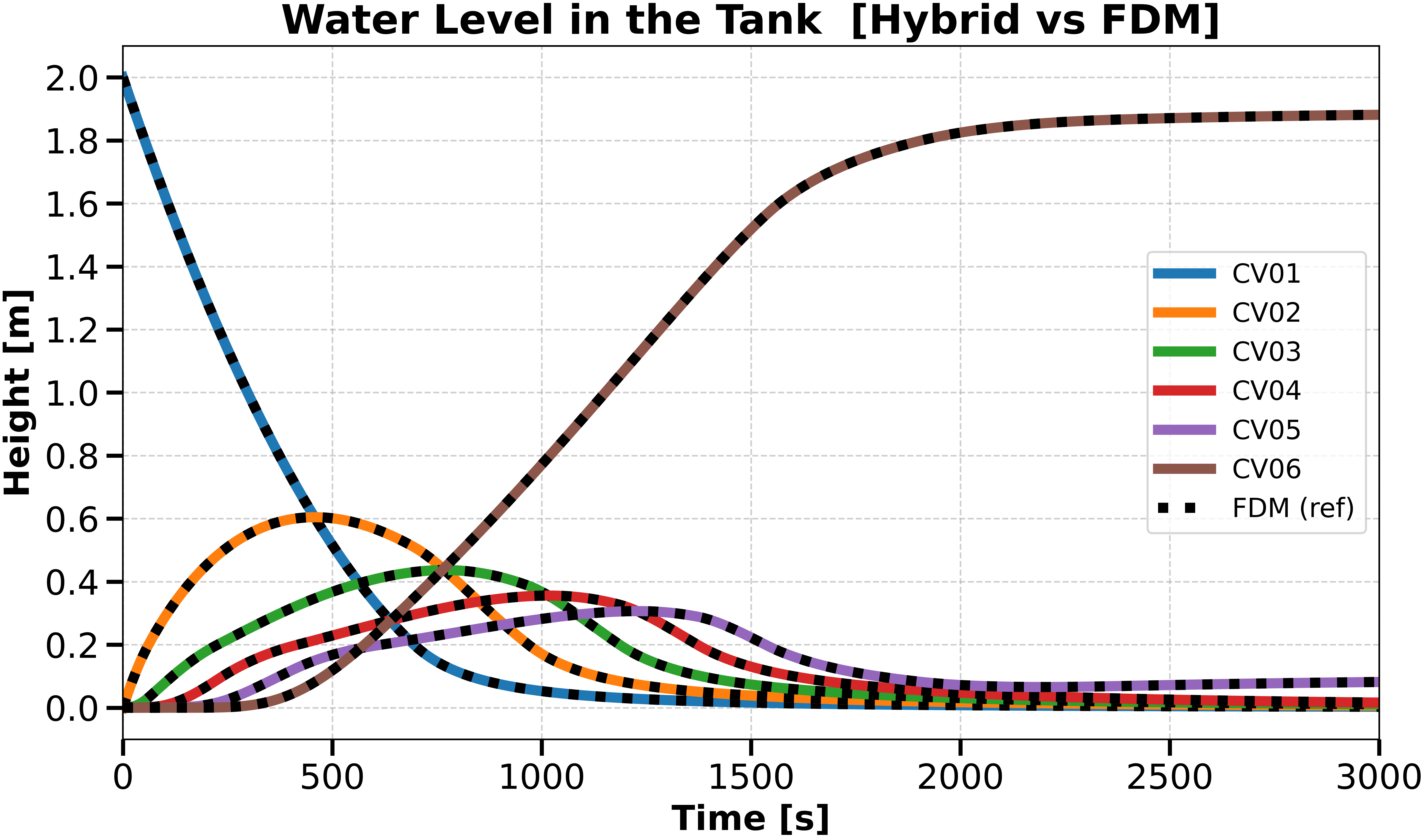}
        \caption{$h_i$, $\Delta t = 0.2$~s}
        \label{fig:h_dt02}
    \end{subfigure}
    \hfill
    \begin{subfigure}[b]{0.48\textwidth}
        \centering
        \includegraphics[width=\textwidth]{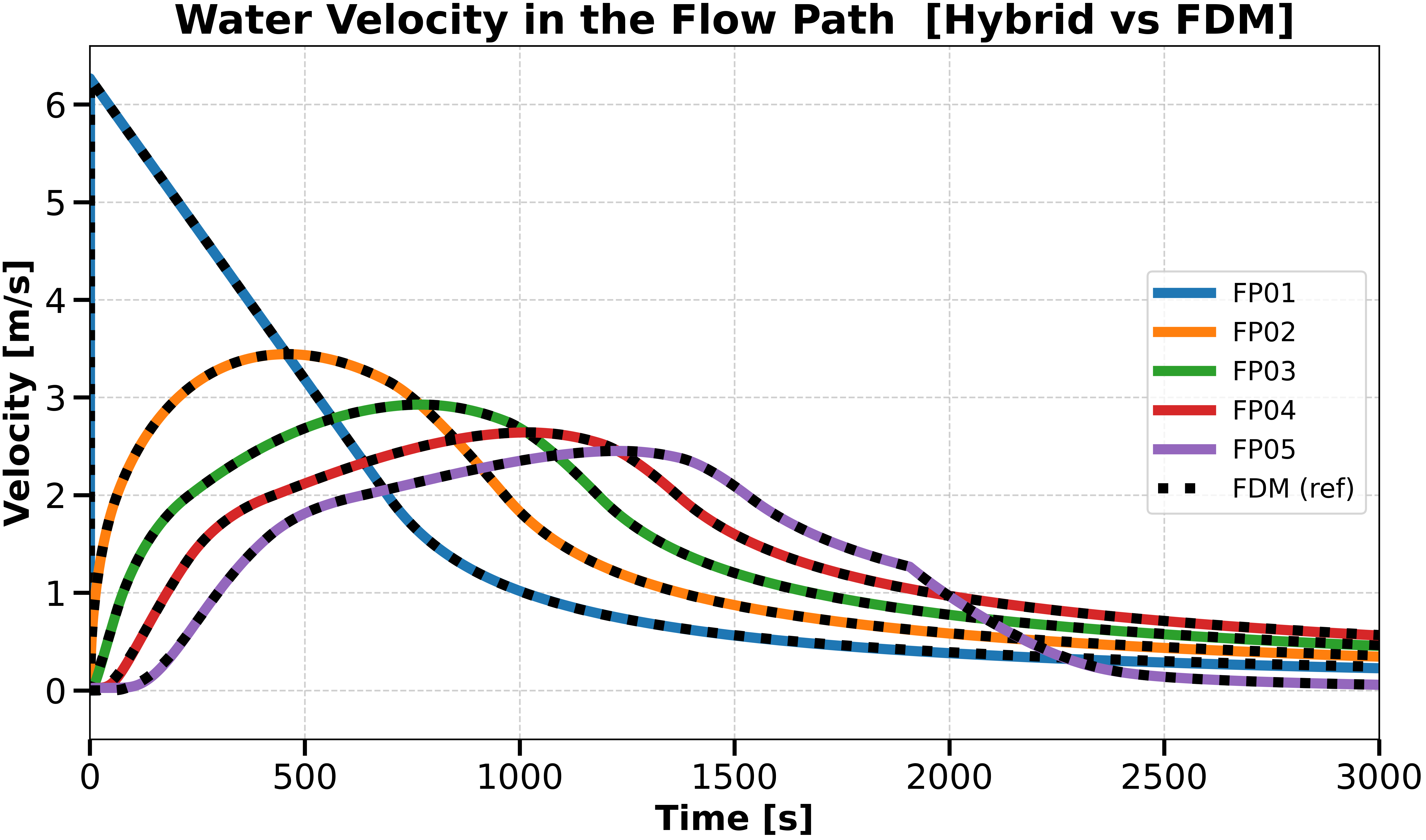}
        \caption{$v_j$, $\Delta t = 0.2$~s}
        \label{fig:v_dt02}
    \end{subfigure}
 
    \vspace{0.3cm}
 
    \begin{subfigure}[b]{0.48\textwidth}
        \centering
        \includegraphics[width=\textwidth]{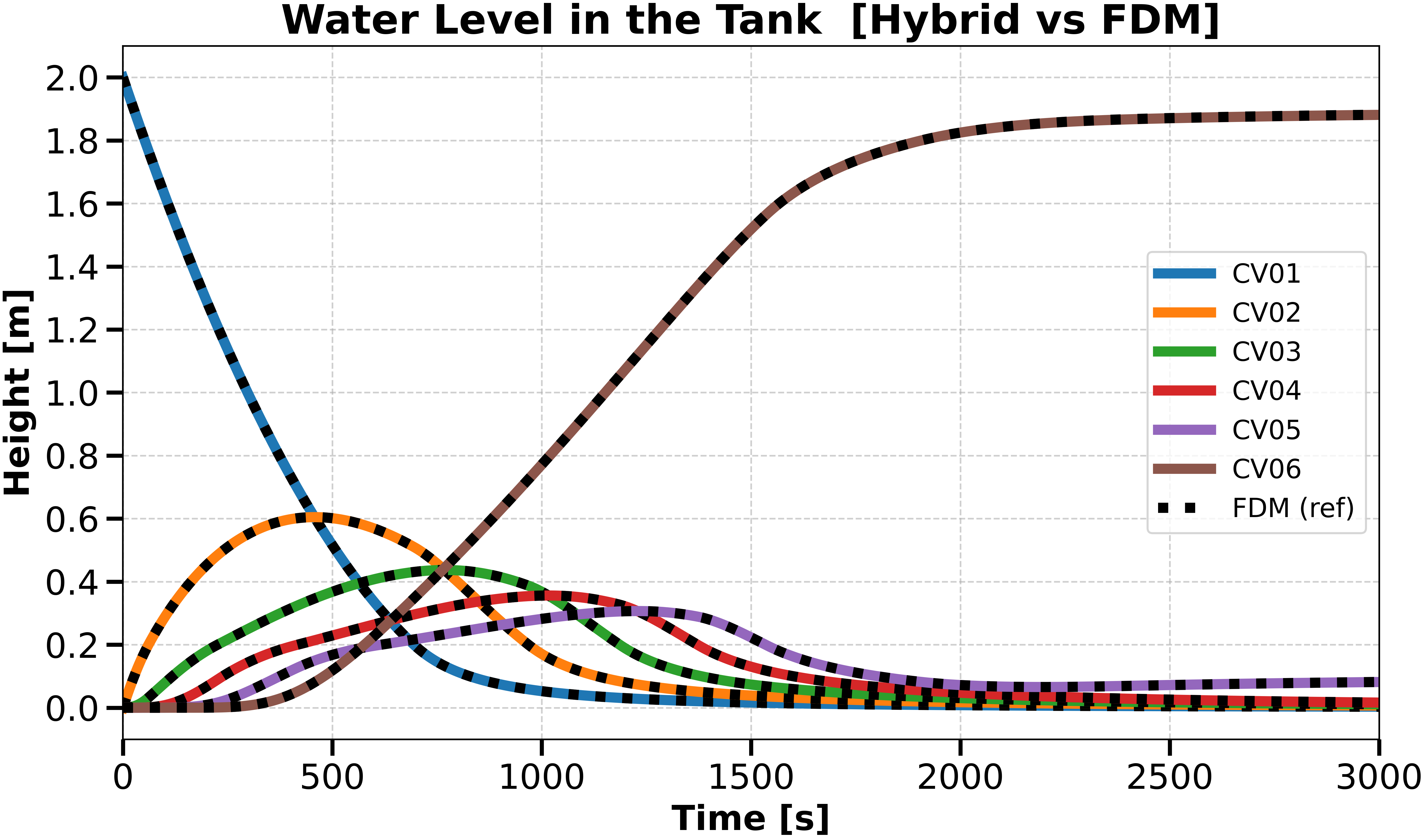}
        \caption{$h_i$, $\Delta t = 0.5$~s}
        \label{fig:h_dt05}
    \end{subfigure}
    \hfill
    \begin{subfigure}[b]{0.48\textwidth}
        \centering
        \includegraphics[width=\textwidth]{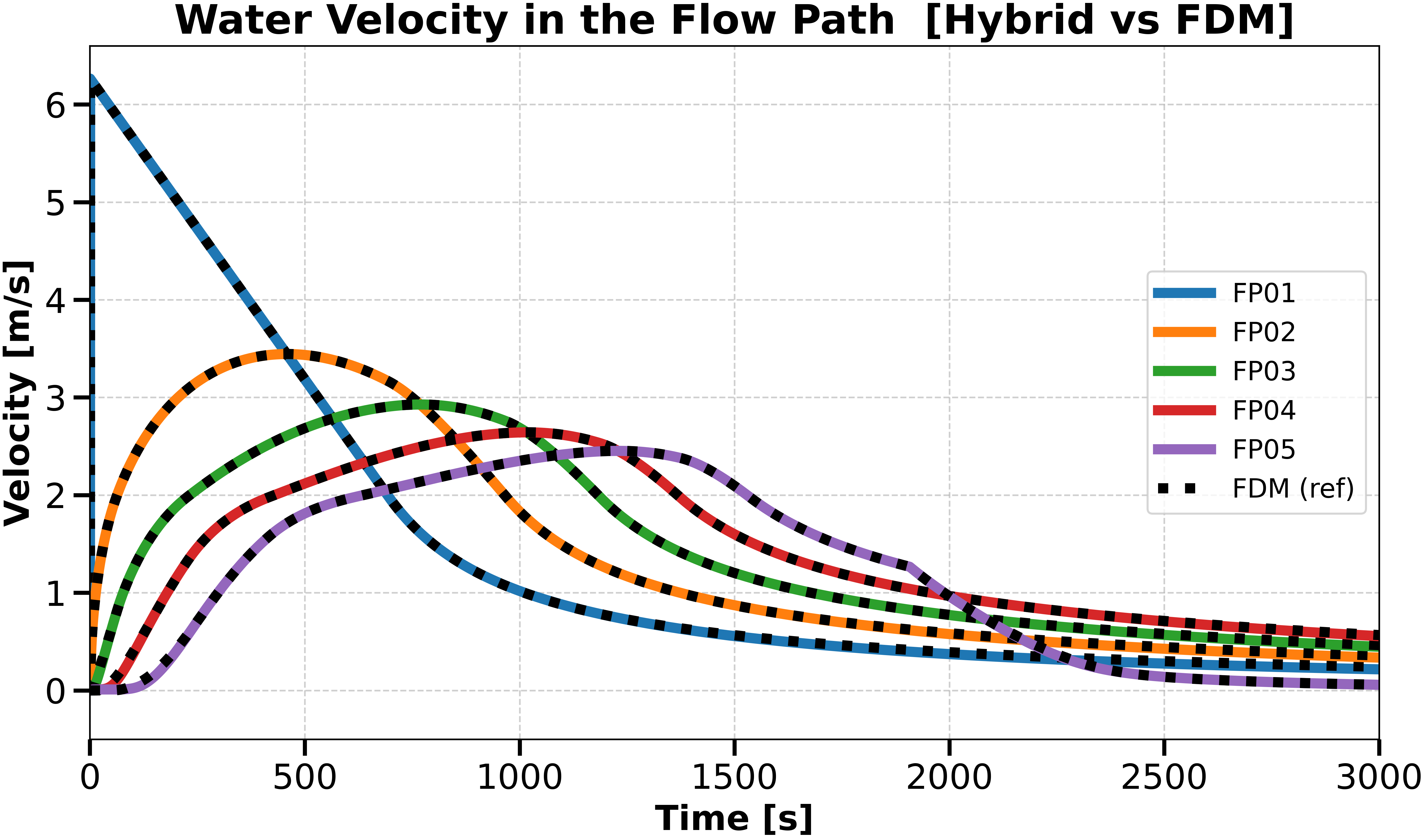}
        \caption{$v_j$, $\Delta t = 0.5$~s}
        \label{fig:v_dt05}
    \end{subfigure}
 
    \vspace{0.3cm}
 
    \begin{subfigure}[b]{0.48\textwidth}
        \centering
        \includegraphics[width=\textwidth]{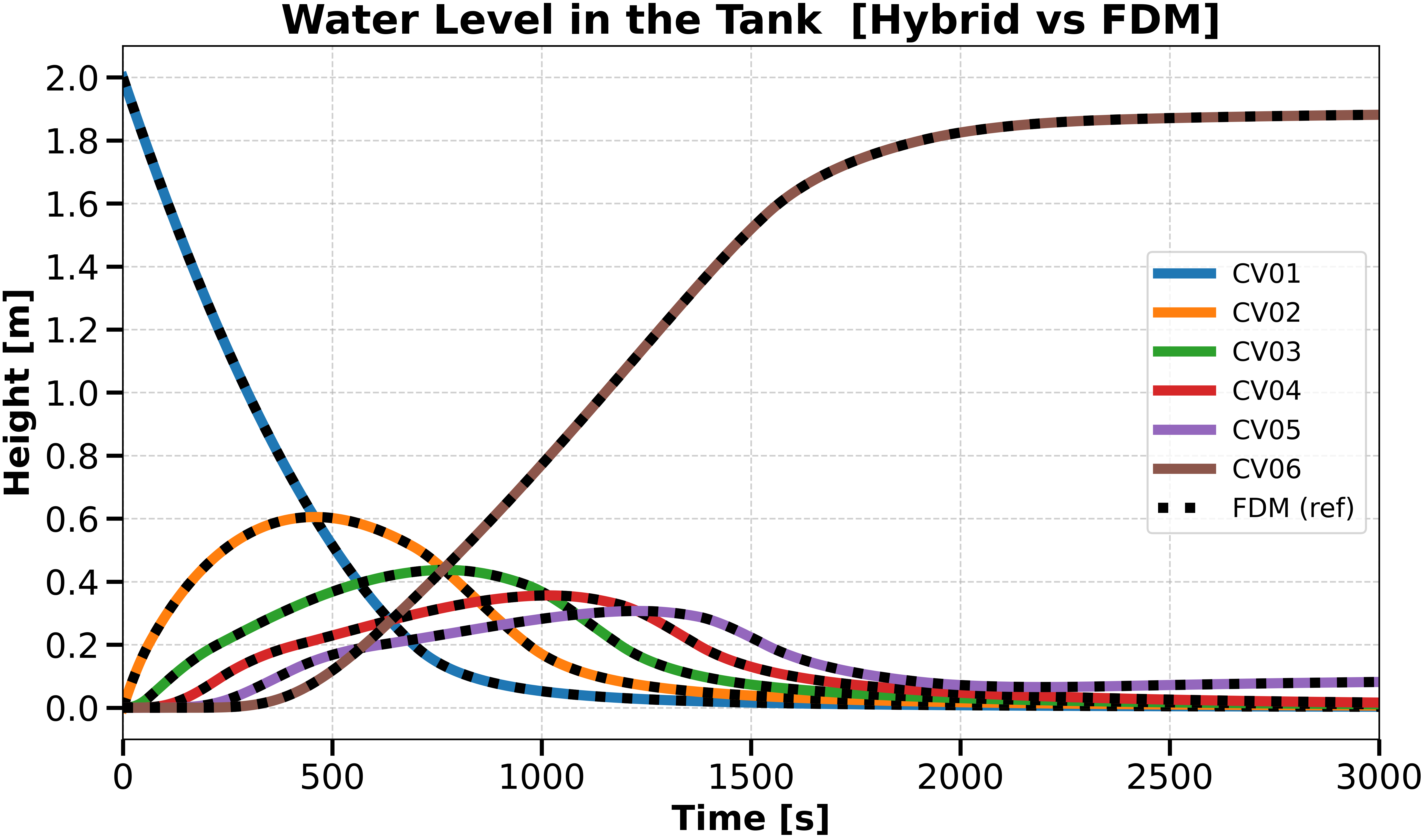}
        \caption{$h_i$, $\Delta t = 1.0$~s}
        \label{fig:h_dt10}
    \end{subfigure}
    \hfill
    \begin{subfigure}[b]{0.48\textwidth}
        \centering
        \includegraphics[width=\textwidth]{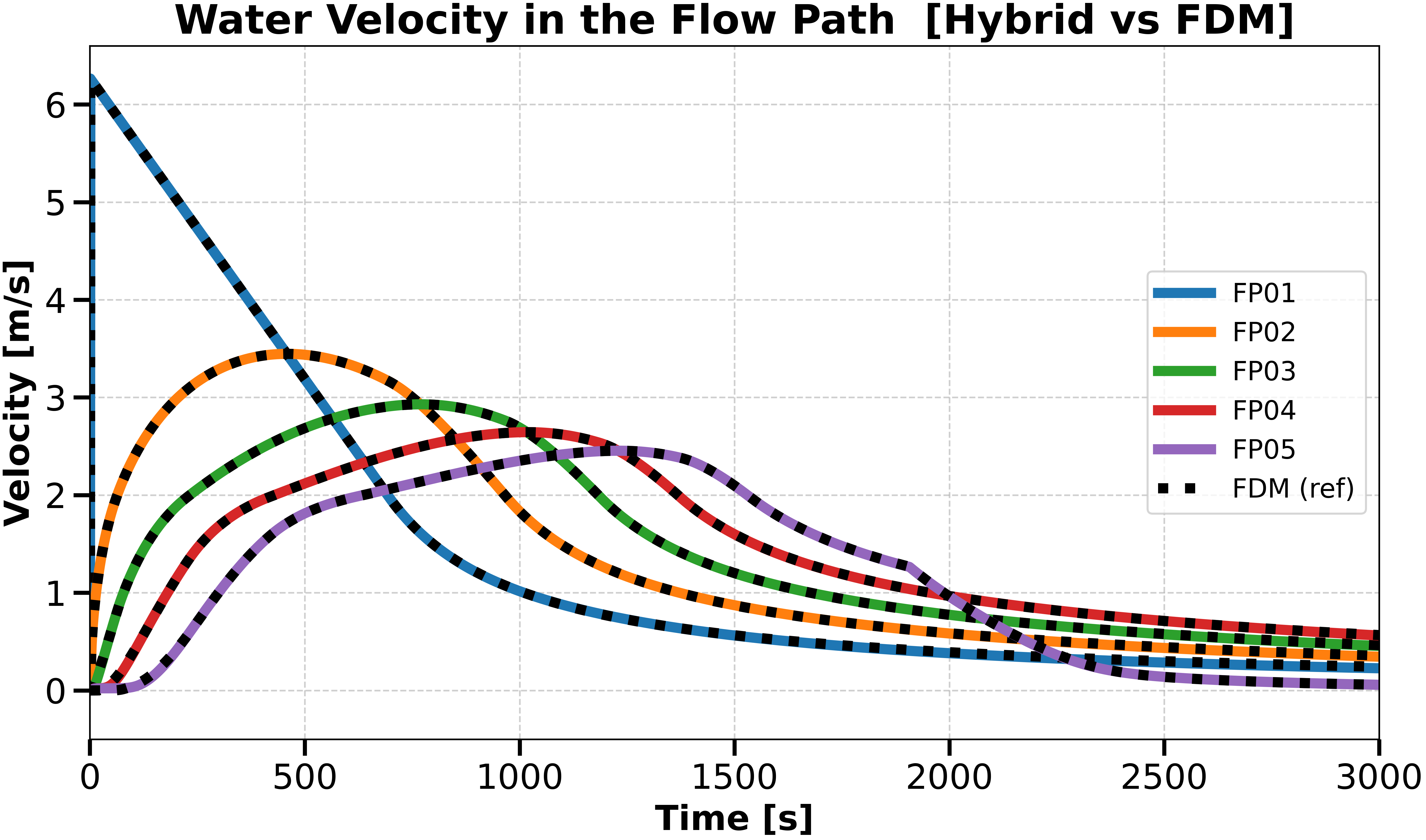}
        \caption{$v_j$, $\Delta t = 1.0$~s}
        \label{fig:v_dt10}
    \end{subfigure}
 
    \caption{Verification under the nominal initial condition ($H_\mathrm{init} = [2,\;0,\;0,\;0,\;0,\;0]$~m).
    Each row corresponds to a time step ($\Delta t = 0.2$, $0.5$, $1.0$~s from top to bottom).
    Left column: CV water levels; right column: FP velocities.
    Solid: hybrid PINN--FDM; dashed: reference FDM.}
    \label{fig:verification_dt}
\end{figure}

Figure~\ref{fig:verification_dt} presents the results, where each row corresponds to a time step ($\Delta t = 0.2$, $0.5$, $1.0$~s) and the two columns show the CV water level histories (left) and FP velocity profiles (right), respectively.
At all three time steps, the hybrid framework closely reproduces the 
reference FDM solution for both CV water levels and FP velocities, 
capturing the cascading transient dynamics described in 
Section~\ref{sec:3.1} with consistent accuracy.
 
Table~\ref{tab:dt_errors} quantifies the prediction errors.
The water level MAE remains on the order of $10^{-5}$~m across all three time steps, with the MSE on the order of $10^{-8}$--$10^{-9}$~m$^2$.
The velocity MAE stays within $3$--$6 \times 10^{-3}$~m/s, with the MSE spanning $3.20 \times 10^{-5}$--$1.02 \times 10^{-4}$~(m/s)$^2$.
Notably, the errors do not increase monotonically with $\Delta t$; the $\Delta t = 1.0$~s case yields errors comparable to those at $\Delta t = 0.2$~s, confirming that the framework maintains consistent accuracy over the tested range.
 
\begin{table}[H]
    \centering
    \caption{Prediction errors under the nominal initial condition for three time steps.
    Errors are computed over all CVs (water level) and all FLs (velocity).}
    \label{tab:dt_errors}
    \begin{tabular}{l cc cc}
        \toprule
        & \multicolumn{2}{c}{Water level $h$ (m)}
        & \multicolumn{2}{c}{Velocity $v$ (m/s)} \\
        \cmidrule(lr){2-3} \cmidrule(lr){4-5}
        $\Delta t$ (s) & MAE & MSE & MAE & MSE \\
        \midrule
        0.2  & $9.32\times10^{-5}$   & $1.78\times10^{-8}$   & $3.08\times10^{-3}$   & $3.20\times10^{-5}$   \\
        0.5  & $1.01\times10^{-4}$   & $2.26\times10^{-8}$   & $5.55\times10^{-3}$   & $1.02\times10^{-4}$   \\
        1.0  & $7.85\times10^{-5}$   & $9.92\times10^{-9}$   & $3.21\times10^{-3}$   & $3.59\times10^{-5}$   \\
        \bottomrule
    \end{tabular}
\end{table}

Table~\ref{tab:cost_comparison} compares the wall-clock times of the two 
approaches. Under the present simplified governing equations, the hybrid 
PINN--FDM framework is approximately 25$\times$ slower than the reference FDM solver across all three time steps. Although the open-tank assumption eliminates the inter-CV pressure coupling and thereby removes the need to assemble and invert a global velocity matrix, the reference FDM solver still performs iterative linearization of the nonlinear friction term $|v|\,v$ at each FP within every time step. The observed speed disadvantage of the hybrid framework is therefore attributable to the overhead of the PINN forward pass (floating-point inference, and host--device data transfer), which exceeds the cost of 
the per-FP iterative friction solve in the current configuration. It is worth noting, however, that the PINN inference cost is fixed at a  single forward pass per FP per time step regardless of the complexity of 
the governing equations, whereas the cost of the conventional iterative 
solver grows with the degree of nonlinearity and inter-equation coupling.

\begin{table}[H]
    \centering
    \caption{Computational cost comparison between the reference FDM solver and the hybrid PINN--FDM framework under the nominal initial condition.}
    \label{tab:cost_comparison}
    \begin{tabular}{l ccc}
        \toprule
        & $\Delta t = 0.2$~s & $\Delta t = 0.5$~s & $\Delta t = 1.0$~s \\
        \midrule
        \multicolumn{4}{l}{\textit{Reference FDM solver}} \\
        \quad time (s)        & 0.179  & 0.075  & 0.036  \\
        \midrule
        \multicolumn{4}{l}{\textit{Hybrid PINN--FDM framework}} \\
        \quad Total inference (s)         & 5.051  & 2.043  & 1.007  \\
        \midrule
        Speedup ratio                     & 0.04x  & 0.04x  & 0.04x  \\
        \bottomrule
    \end{tabular}
\end{table}
 
\subsection{Generalization across Initial Conditions}
\label{sec:4.3}
 
To demonstrate that the hybrid framework generalizes beyond the nominal configuration, it is applied to five test cases with diverse initial water level distributions, listed in the first two columns of Table~\ref{tab:ic_errors}.
The cases probe different regions of the trained parameter space:
Case~1 fills two upstream CVs with unequal levels;
Cases~2 and 5 distribute water across three or four CVs with unequal levels;
Case~3 fills two adjacent CVs with a moderate head difference;
and Case~4 distributes equal water levels across four CVs, resulting in zero initial driving heads across FL01--FL03.
All simulations use $\Delta t = 1.0$~s.
 
\begin{figure}[H]
    \centering
    \begin{subfigure}[b]{0.48\textwidth}
        \centering
        \includegraphics[width=\textwidth]{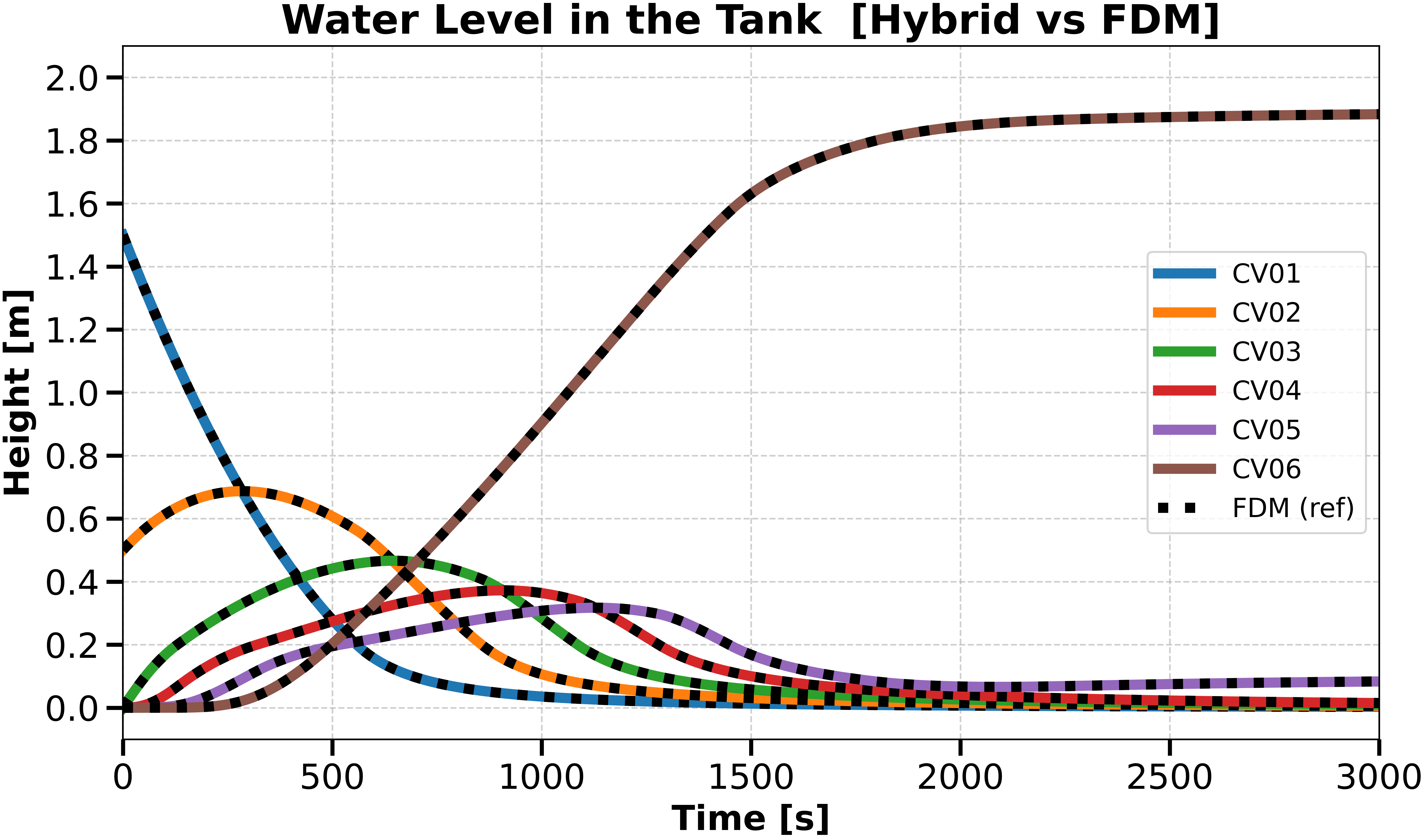}
        \caption{$h_i$, Case~1}
        \label{fig:ic1_h}
    \end{subfigure}
    \hfill
    \begin{subfigure}[b]{0.48\textwidth}
        \centering
        \includegraphics[width=\textwidth]{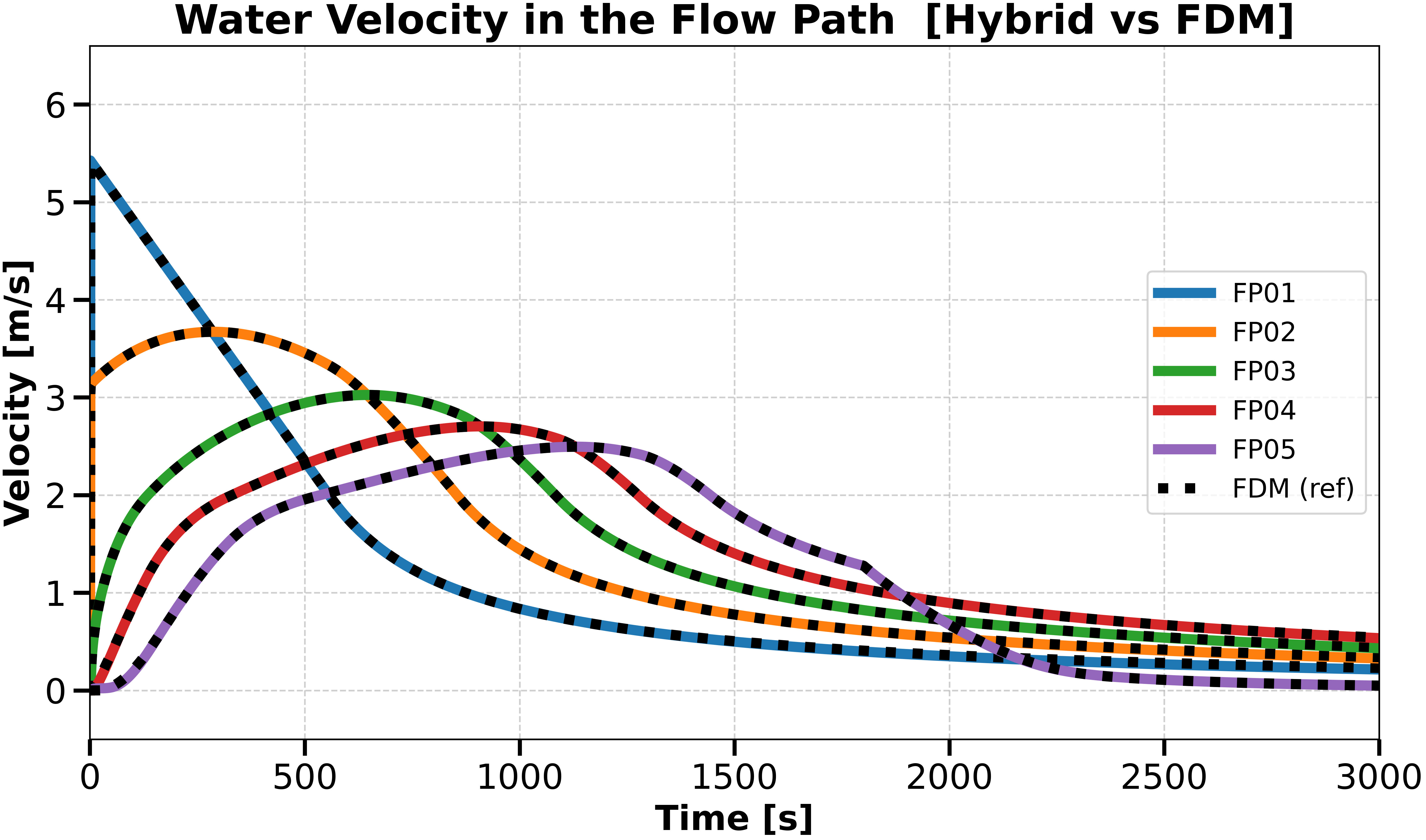}
        \caption{$v_j$, Case~1}
        \label{fig:ic1_v}
    \end{subfigure}
 
    \vspace{0.3cm}
 
    \begin{subfigure}[b]{0.48\textwidth}
        \centering
        \includegraphics[width=\textwidth]{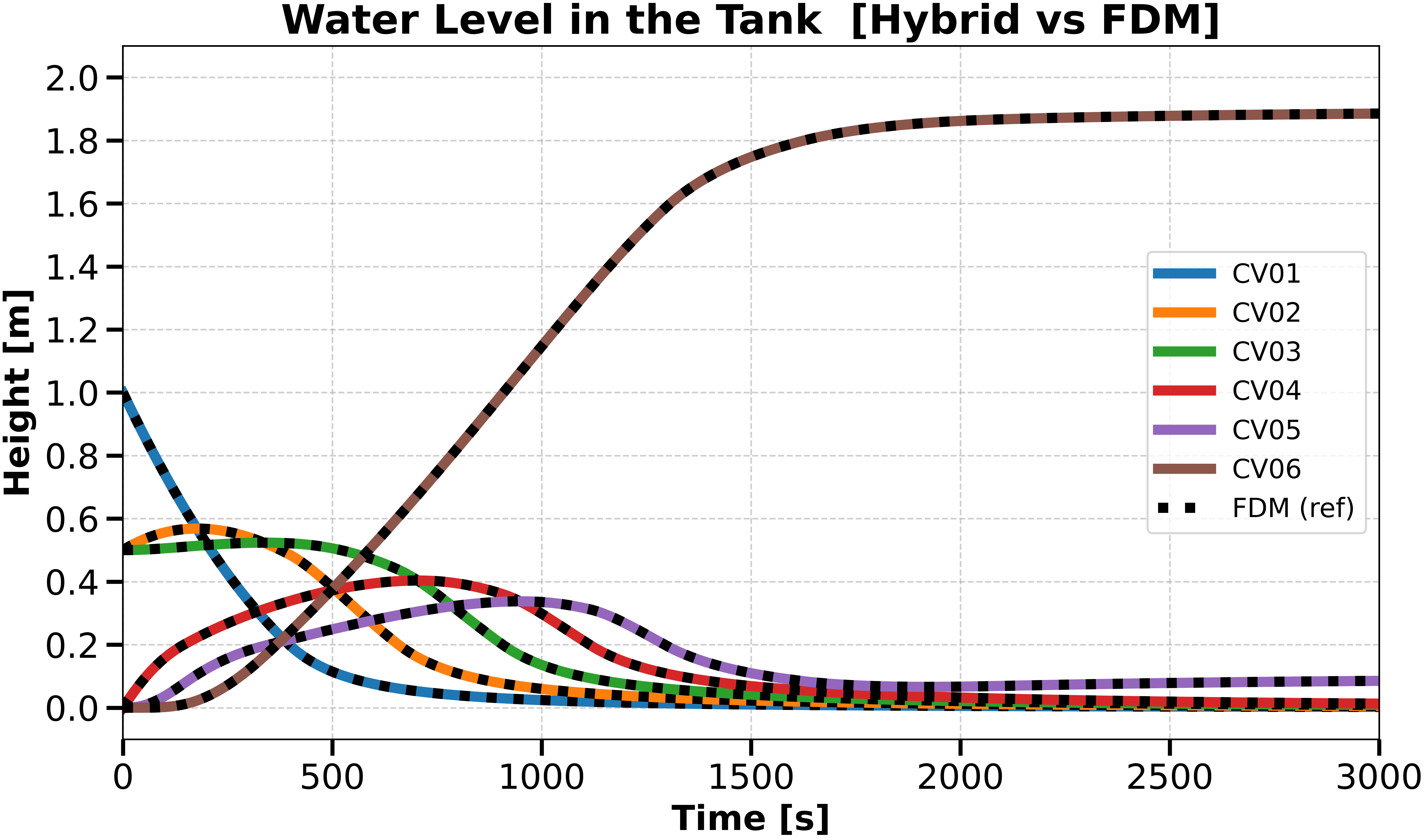}
        \caption{$h_i$, Case~2}
        \label{fig:ic2_h}
    \end{subfigure}
    \hfill
    \begin{subfigure}[b]{0.48\textwidth}
        \centering
        \includegraphics[width=\textwidth]{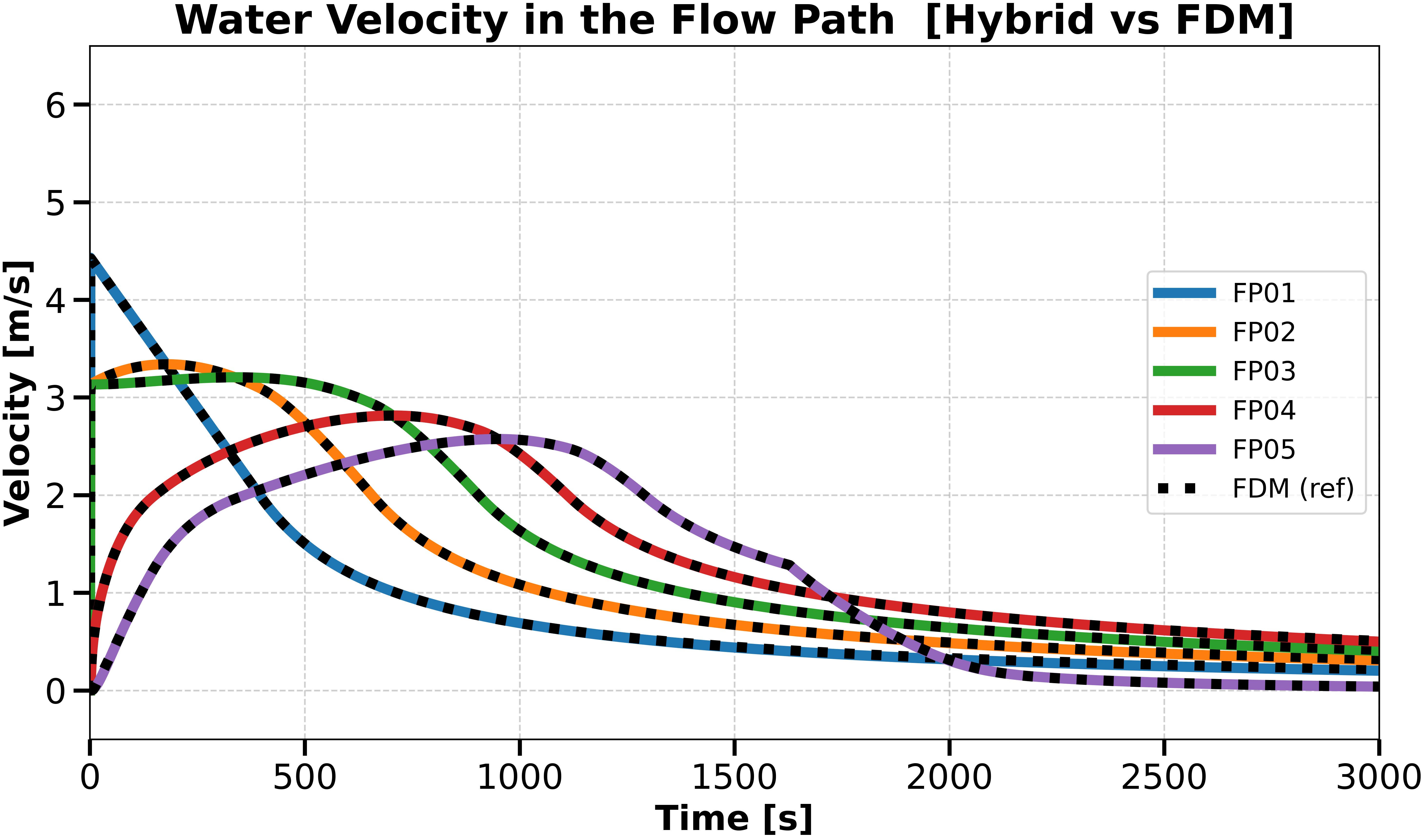}
        \caption{$v_j$, Case~2}
        \label{fig:ic2_v}
    \end{subfigure}
 
    \caption{Generalization to unseen initial conditions ($\Delta t = 1.0$~s).
    Top row: Case~1 ($H_\mathrm{init} = [1.5,\;0.5,\;0,\;0,\;0,\;0]$~m);
    bottom row: Case~2 ($H_\mathrm{init} = [1.0,\;0.5,\;0.5,\;0,\;0,\;0]$~m).
    Left: CV water levels; right: FP velocities.
    Solid: hybrid PINN--FDM; dashed: reference FDM.}
    \label{fig:ic_generalization}
\end{figure}

Figure~\ref{fig:ic_generalization} presents time histories for two representative cases that illustrate qualitatively different transient behaviors.
In Case~1 (Figs.~\ref{fig:ic1_h} and \ref{fig:ic1_v}, top row; $H_\mathrm{init} = [1.5,\;0.5,\;0,\;0,\;0,\;0]$~m), water is present in both CV01 and CV02 from the outset, activating FL01 and FL02 simultaneously and producing overlapping velocity peaks in the early transient.
The hybrid framework captures the simultaneous draining, the water level crossover between CV01 and CV02, and the subsequent cascading to downstream CVs.
In Case~2 (Figs.~\ref{fig:ic2_h} and \ref{fig:ic2_v}, bottom row; $H_\mathrm{init} = [1.0,\;0.5,\;0.5,\;0,\;0,\;0]$~m), the head difference across FL02 is initially zero, so FL02 remains quiescent until inflow from FL01 raises the CV02 level above CV03.
This delayed activation is correctly reproduced, and the resulting more gradual redistribution---with lower peak velocities and a longer time to equilibrium---is accurately tracked.
 
Table~\ref{tab:ic_errors} summarizes the errors across all five cases.
The water level MAE remains consistently on the order of $9 \times 10^{-5}$~m ($9.05$--$9.43 \times 10^{-5}$~m), while the velocity MAE ranges from $3.24 \times 10^{-3}$~m/s (Case~1) to $4.17 \times 10^{-3}$~m/s (Case~4).
Case~4 exhibits the highest velocity error among the five cases, but the increase is modest and the water level accuracy is unaffected.

\begin{table}[H]
    \centering
    \caption{Prediction errors across five initial condition cases ($\Delta t = 1.0$~s).
    Errors are computed over all CVs (water level) and all FLs (velocity).}
    \label{tab:ic_errors}

    \resizebox{\linewidth}{!}{
    \begin{tabular}{l l cc cc}
        \toprule
        & & \multicolumn{2}{c}{Water level $h$ (m)}
        & \multicolumn{2}{c}{Velocity $v$ (m/s)} \\
        \cmidrule(lr){3-4} \cmidrule(lr){5-6}
        Case & $H_\mathrm{init}$ (m) & MAE & MSE & MAE & MSE \\
        \midrule
        1 & $[1.5,\; 0.5,\; 0,\; 0,\; 0,\; 0]$             & $9.43\times10^{-5}$   & $1.77\times10^{-8}$   & $3.24\times10^{-3}$   & $3.34\times10^{-5}$   \\
        2 & $[1.0,\; 0.5,\; 0.5,\; 0,\; 0,\; 0]$           & $9.35\times10^{-5}$   & $1.64\times10^{-8}$   & $3.57\times10^{-3}$   & $3.70\times10^{-5}$   \\
        3 & $[1.3,\; 0.7,\; 0,\; 0,\; 0,\; 0]$             & $9.32\times10^{-5}$   & $1.71\times10^{-8}$   & $3.37\times10^{-3}$   & $3.51\times10^{-5}$   \\
        4 & $[0.5,\; 0.5,\; 0.5,\; 0.5,\; 0,\; 0]$         & $9.29\times10^{-5}$   & $1.49\times10^{-8}$   & $4.17\times10^{-3}$   & $4.54\times10^{-5}$   \\
        5 & $[1.0,\; 0.5,\; 0.3,\; 0.2,\; 0,\; 0]$         & $9.05\times10^{-5}$   & $1.50\times10^{-8}$   & $3.50\times10^{-3}$   & $3.64\times10^{-5}$   \\
        \bottomrule
    \end{tabular}
    }
\end{table}
 
The overall uniformity of error magnitudes confirms that the framework generalizes across the trained parameter range.
Taken together with the time-step robustness demonstrated in Section~\ref{sec:4.2}, the hybrid framework achieves water level MAE of $\mathcal{O}(10^{-5})$~m and velocity MAE of $\mathcal{O}(10^{-3})$~m/s across all tested configurations, maintains consistent accuracy for $\Delta t$ ranging from $0.2$ to $1.0$~s without retraining, and generalizes across five distinct initial conditions.
These results demonstrate that the P2F framework provides a practical data-free surrogate for thermal-hydraulic system codes.

\section{Conclusion}\label{sec:6}

This study presented the Parameterized PINNs coupled with FDM (P2F) method, a node-assigned hybrid framework for thermal-hydraulic system simulation, addressing two gaps identified in the literature: the absence of a data-free surrogate for nuclear system codes and the lack of an node-assigned hybrid coupling between a data-free PINN and a conventional numerical solver.
 
The framework was built upon a parameterized NA-PINN that accepts the water-level difference $\Delta h$, the initial velocity $v_o$, and the time coordinate $t$ as inputs, enabling a single trained network to serve as a parameterized surrogate for the momentum conservation equation across all flow paths.
By embedding the initial condition directly into the network output through a hard constraint formulation, the training loss reduces to a single physics-based residual term, eliminating the need for multi-objective balancing.
Standalone verification against reference FDM solutions confirmed that the parameterized PINN accurately reproduces the momentum equation dynamics across the training parameter space, with MAE of $\mathcal{O}(10^{-3})$~m/s.
 
The trained PINN was then coupled with an FDM solver in the P2F framework: the PINN handles the nonlinear momentum equation via a single forward pass, while the FDM solver advances the mass conservation equation to enforce discrete mass conservation at every time step.
Verification on a six-tank gravity-driven draining scenario demonstrated water level MAE of $\mathcal{O}(10^{-5})$~m and velocity MAE of $\mathcal{O}(10^{-3})$~m/s across all tested configurations.
The framework maintained consistent accuracy for time steps ranging from 0.2 to 1.0~s without retraining and generalized across five distinct initial conditions, confirming its robustness and parametric flexibility.
 
These results establish the P2F method as the first data-free surrogate approach for a nuclear thermal-hydraulic system code that requires no simulation data for training and no retraining for new initial conditions, while remaining directly compatible with the FDM-based structure of existing system codes such as MELCOR.

Several avenues for future work remain.
First, the present study considers an idealized scenario under open-tank conditions where the fixed flow direction permits a sequential upwind-scheme coupling.
We plan to extend the framework to more realistic severe accident scenarios, including closed systems with non-negligible pressure differentials, bidirectional flow, and the corresponding matrix-based implicit coupling consistent with MELCOR's numerical scheme.
Second, the current implementation is limited to the CVH/FP module; incorporating other key MELCOR packages---such as the Heat Structure (HS) and Radionuclide (RN) modules---would enable a more comprehensive multi-physics simulation framework and further demonstrate the scalability of the P2F approach.
Third, the present hybrid framework is approximately 25$\times$ slower 
than the reference FDM solver under the simplified open-tank equations; 
systematically benchmarking the computational cost under progressively 
more complex governing equations---including closed-system configurations with inter-CV pressure coupling and global velocity matrix assembly---is necessary to identify the regime in which the hybrid approach becomes cost-competitive.

\section*{Acknowledgments}
This work was supported by the Ministry of Science and ICT of Korea (No. RS-2024-00355857) and by the Nuclear Safety Research Program through the Korea Foundation of Nuclear Safety (KoFONS), funded by the Nuclear Safety and Security Commission (NSSC) of the Republic of Korea (No. RS-2024-00403364).


\bibliographystyle{elsarticle-num} 
\bibliography{sample}






\end{document}